\documentclass{article}

\usepackage[final]{neurips_2021}
\usepackage{graphicx}
\usepackage{subcaption} 

\usepackage{enumitem,amssymb}
\newlist{todolist}{itemize}{2}
\setlist[todolist]{label=$\square$}
\usepackage{pifont}
\usepackage[utf8]{inputenc} 
\usepackage[T1]{fontenc}    
\usepackage{hyperref}       
\usepackage{url}            
\usepackage{booktabs}       
\usepackage{amsfonts}       
\usepackage{nicefrac}       
\usepackage{microtype}      
\usepackage{xcolor}         
\usepackage{bbm}         
\usepackage{comment}

\usepackage{amsmath}

\newtheorem{lemma}{Lemma}
\newtheorem{proposition}{Proposition}
\newenvironment{customlemma}[1]
  {\renewcommand\thelemma{#1}\lemma}
  {\endlemma}
\newenvironment{customproposition}[1]
  {\renewcommand\theproposition{#1}\proposition}
  {\endproposition}

\usepackage{xcolor}
\definecolor{dark-red}{rgb}{0.4,0.15,0.15}
\definecolor{dark-blue}{rgb}{0.15,0.15,0.4}
\definecolor{medium-blue}{rgb}{0,0,0.5}
\hypersetup{
   colorlinks, linkcolor={dark-blue},
   citecolor={dark-blue}, urlcolor={medium-blue}
}

\title{Dangers of Bayesian Model Averaging \\ under Covariate Shift}

\author{%
  Pavel Izmailov \\
  NYU
  \And
  Patrick Nicholson \\
  Covera Health
  \And
  Sanae Lotfi \\
  NYU
  \And
  Andrew Gordon Wilson \\
  NYU
}
  
\begin{document}
\normalsize

\maketitle

\begin{abstract}
  
  Approximate Bayesian inference for neural networks is considered a robust alternative to standard training, often providing good performance on out-of-distribution data. However, 
  Bayesian neural networks (BNNs) with high-fidelity approximate inference via full-batch Hamiltonian Monte Carlo achieve poor generalization under covariate shift, even underperforming classical estimation. We explain this surprising result, showing how a Bayesian model average can in fact be problematic under covariate shift, particularly in cases where linear dependencies in the input features cause a lack of posterior contraction. We additionally show why the same issue does not affect many approximate inference procedures, or classical maximum a-posteriori (MAP) training. Finally, we propose novel priors that improve the robustness of BNNs to many sources of covariate shift.
  
\end{abstract}

\section{Introduction}
\label{sec:intro}

The predictive distributions of deep neural networks are often deployed in critical applications such as medical diagnosis \citep{gulshan2016development, esteva2017dermatologist, filos2019systematic}, and autonomous driving \citep{bojarski2016end, al2017learning, michelmore2020uncertainty}. These applications typically involve \emph{covariate shift}, where the target data distribution is different from the distribution used for training \citep{hendrycks2019benchmarking, arjovsky2021out}. Accurately reflecting uncertainty is crucial for robustness to these shifts \citep{ovadia2019can, roy2021does}. Since Bayesian methods provide a principled approach to representing model (epistemic) uncertainty, they are commonly benchmarked on out-of-distribution (OOD) generalization tasks \citep[e.g.,][]{kendall2017uncertainties, ovadia2019can, chang2019ensemble, dusenberry2020efficient, wilson2020bayesian}.

However, \citet{izmailov2021bayesian} recently showed that Bayesian neural networks (BNNs) with high fidelity inference through Hamiltonian Monte Carlo (HMC) provide shockingly poor OOD generalization performance, despite the popularity and success of approximate Bayesian inference in this setting \citep{gal2016dropout, lakshminarayanan2017simple, ovadia2019can, maddox2019simple, wilson2020bayesian, dusenberry2020efficient, benton2021loss}.

\begin{figure}
\centering
    \begin{tabular}{ccc}
        \hspace{-0.2cm}\includegraphics[height=0.181\textwidth]{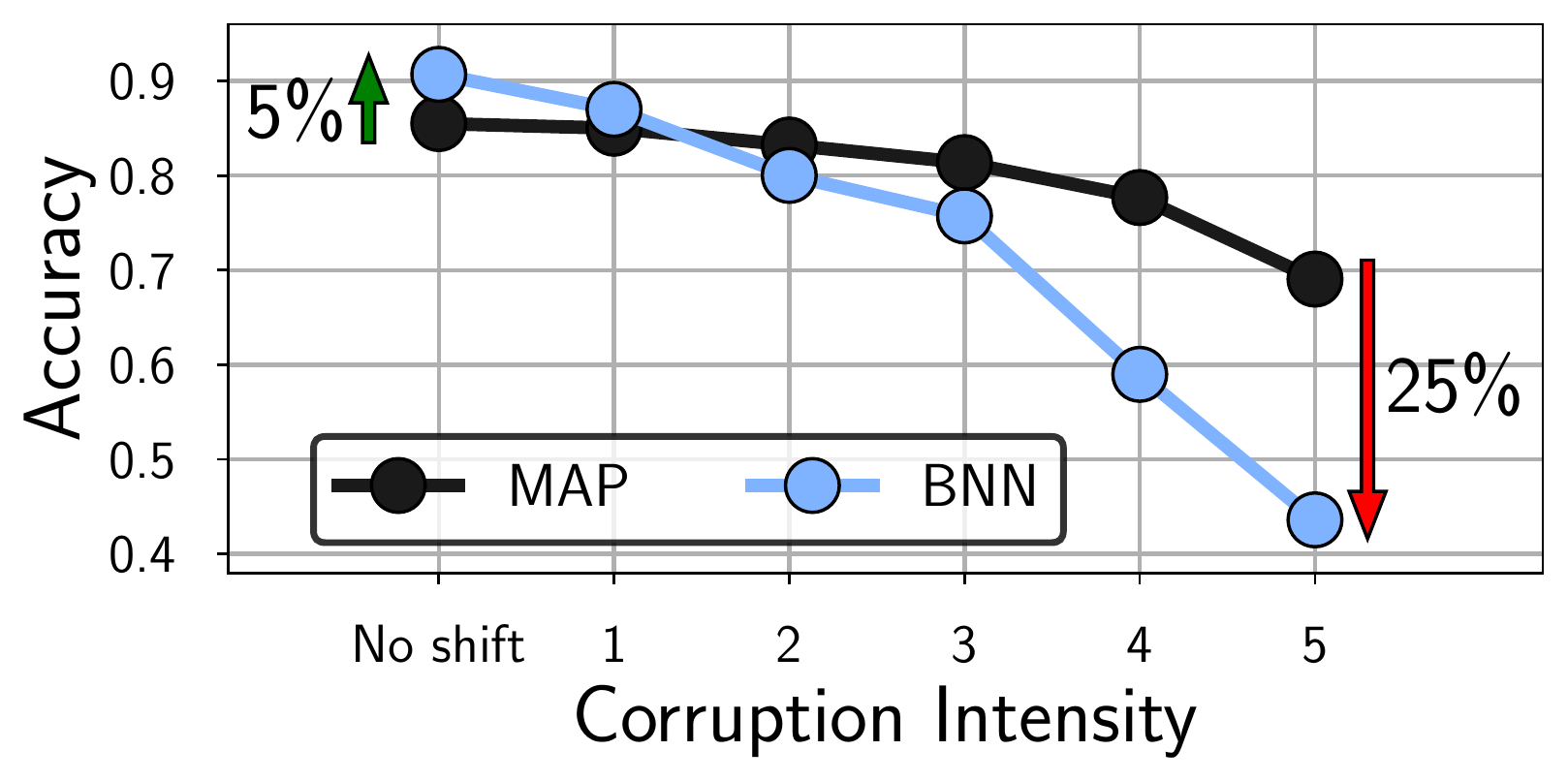}
        &
        \hspace{-0.1cm}\includegraphics[height=0.181\textwidth, trim=0cm -0.cm 0.cm 0cm, clip]{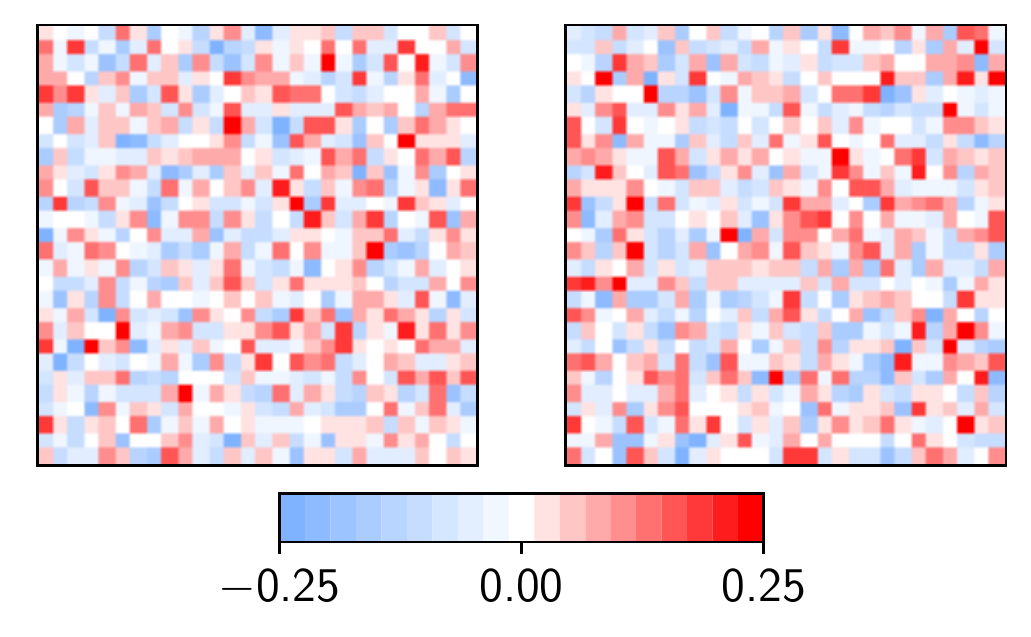}
        &
        \hspace{-0.1cm}\includegraphics[height=0.181\textwidth, trim=0cm -0.cm 0.cm 0cm, clip]{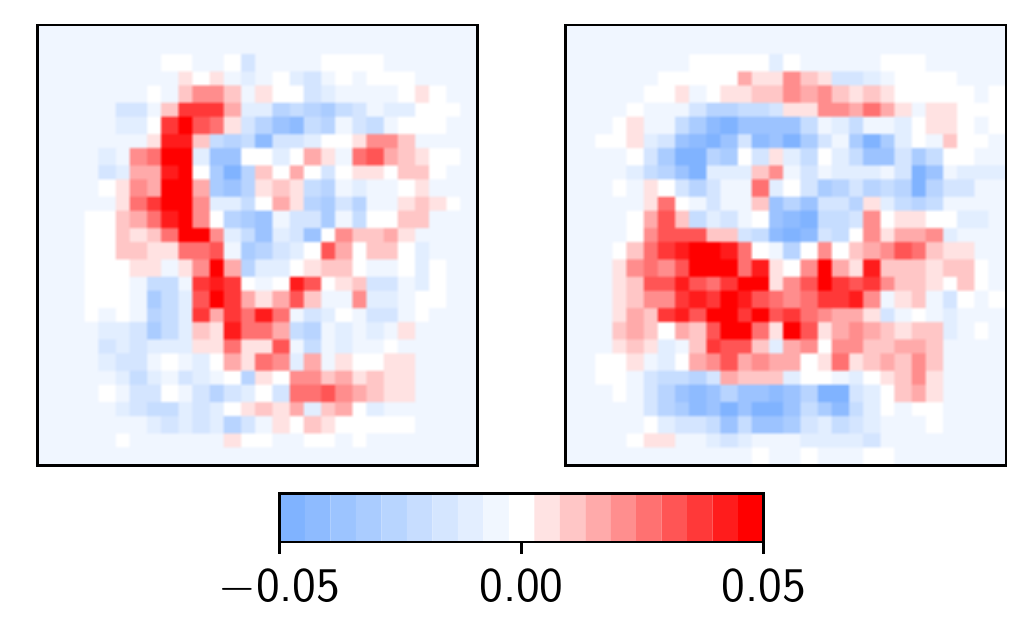}
      \\
        \hspace{-0.2cm}{\small (a) ResNet-20, CIFAR-10-C} &
        \hspace{-0.1cm}{\small (b) BNN weights} &  
        \hspace{-0.1cm}{\small (c) MAP weights}
    \end{tabular}
\caption{\textbf{Bayesian neural networks under covariate shift}.
\textbf{(a)}: Performance of a ResNet-20 on the \textit{pixelate} corruption in CIFAR-10-C.
For the highest degree of corruption, a Bayesian model average underperforms a MAP solution by 25\% (44\% against 69\%) accuracy. 
See \citet{izmailov2021bayesian} for details.
\textbf{(b)}: Visualization of the weights in the first layer of a Bayesian fully-connected network on MNIST sampled via HMC.
\textbf{(c)}: The corresponding MAP weights.
We visualize the weights connecting the input pixels to a neuron in the hidden layer as a $28 \times 28$ image, where each weight is shown
in the location of the input pixel it interacts with.
}
\label{fig:intro_figure}
\end{figure}

In this paper, we seek to understand, further demonstrate, and help remedy this concerning behaviour. 
We show that Bayesian neural networks perform poorly for different types of covariate shift, namely test data corruption, domain shift, and spurious correlations. In \autoref{fig:intro_figure}(a) we see that a ResNet-20 BNN approximated with HMC underperforms a maximum a-posteriori (MAP) solution by $25\%$ on the \textit{pixelate}-corrupted CIFAR-10 test set. 
This result is particularly surprising given that on the in-distribution test data, the BNN outperforms the MAP solution by over $5\%$. 

Intuitively, we find that Bayesian model averaging (BMA) can be problematic under covariate shift as follows. Due to linear dependencies in the features (inputs) of the training data distribution, model parameters corresponding to these dependencies do not affect the predictions on the training data. 
As an illustrative special case of this general setting, consider MNIST digits, which always have black corner pixels (\emph{dead pixels}, with intensity zero). The corresponding first layer weights are always multiplied by zero and have no effect on the likelihood. Consequently, these weights are simply sampled from the prior. If at test time the corner pixels are not black, e.g., due to corruption, these pixel values will be multiplied by random weights sampled from the prior, and propagated to the next layer, significantly degrading performance. On the other hand, classical MAP training drives the unrestricted parameters towards zero due to regularization from the prior that penalizes the parameter norm, and will not be similarly affected by noise at test time. Here we see a major difference in robustness between optimizing a posterior for MAP training in comparison to a posterior weighted model average.

As a motivating example, in \autoref{fig:intro_figure}(b, c) we visualize the weights in the first layer of a fully-connected network for a sample from the BNN posterior and the MAP solution on the MNIST dataset.
The MAP solution weights are highly structured, while the BNN sample appears extremely noisy, similar to a draw from the Gaussian prior. In particular the weights corresponding to {dead pixels} (i.e. pixel positions that are black for all the MNIST images) near the boundary of the input image are set near zero (shown in white) by the MAP solution, but sampled randomly by the BNN. If at test time the data is corrupted, e.g. by Gaussian noise, and the pixels near the boundary of the image are activated, the MAP solution will ignore these pixels, while the predictions of the BNN will be significantly affected. 

Dead pixels are a special case of our more general findings: we show that the dramatic lack of robustness for Bayesian neural networks is fundamentally caused by \emph{any} linear dependencies in the data, combined with models that are non-linear in their parameters. Indeed, we consider a wide range of covariate shifts, including domain shifts. These robustness issues have the potential to impact virtually \emph{every} real-world application of Bayesian neural networks, since train and test rarely come from exactly the same distribution.

Based on our understanding, we introduce a novel prior that assigns a low variance to the weights in the first layer corresponding to directions orthogonal to the data manifold, leading to improved generalization under covariate shift. We additionally study the effect of non-zero mean corruptions and accordingly propose a second prior that constrains the sum of the weights, resulting in further improvements in OOD generalization for Bayesian neural networks.

Our code is \href{https://github.com/izmailovpavel/bnn_covariate_shift}{\underline{available here}}.

\section{Background}
\label{sec:background}

\textbf{Bayesian neural networks.}
\quad
A Bayesian neural network model is specified by the prior distribution $p(w)$ over the weights $w$ of the model, and the likelihood function $p(y \vert x, w)$, where $x$ represents the input features and $y$ represents the target value.
Following Bayes' rule, the \textit{posterior} distribution over the parameters $w$ after observing the dataset $\mathcal D = \{(x_i, y_i) \vert i=1, \ldots, n\}$ is given by 
\begin{equation}
    \label{eq:bayes_rule}
    p(w \vert \mathcal D) =
    \frac{p(D \vert w) \cdot p(w)}{\int_{w'} p(D \vert w') \cdot p(w') dw'} =
    \frac{\prod_{i=1}^n p(y_i \vert x_i, w) \cdot p(w)}{\int_{w'} \prod_{i=1}^n p(y_i \vert x_i, w') \cdot p(w') dw'},
\end{equation}
where we assume that the likelihood is independent over the data points.
The posterior in \autoref{eq:bayes_rule} is then used to make predictions for a new input $x$ according to the \textit{Bayesian model average}:
\begin{equation}
    \label{eq:bma}
    p(y \vert x) =
    \int p(y \vert x, w) \cdot p(w \vert \mathcal D) dw.
\end{equation}
Unfortunately, computing the BMA in \autoref{eq:bma} is intractable for Bayesian neural networks.
Hence, a number of approximate inference methods have been developed.
In this work, we use full-batch HMC \citep{neal2011mcmc, izmailov2021bayesian}, as it provides a high-accuracy posterior approximation.
For a more detailed discussion of Bayesian deep learning see \citet{wilson2020bayesian}.

\textbf{Maximum	a-posteriori (MAP) estimation.}
\quad
In contrast with Bayesian model averaging, a MAP estimator uses the single setting of weights (hypothesis) that maximizes the posterior density
 $ w_{\text{MAP}} = \underset{w}{\mathrm{argmax}} \,  p(w \vert \mathcal D) = \underset{w}{\mathrm{argmax}}(\log p(\mathcal{D}|w) + \log p(w))$, where the log prior can be viewed as a regularizer. For example, if we use a Gaussian prior on $w$, then $\log p(w)$ will penalize the $\ell_2$ norm of the parameters, driving parameters that do not improve the log likelihood $\log p(\mathcal{D} \vert w)$ to zero. MAP is the standard approach to training neural networks and our baseline for \emph{classical training} throughout the paper. We perform MAP estimation with SGD unless otherwise specified.

\textbf{Covariate shift.}
\quad
In this paper, we focus on the covariate shift setting.
We assume the training dataset $\mathcal D_{\text{train}}$ consists of i.i.d. samples from the distribution $p_\text{train}(x, y) = p_\text{train}(x) \cdot p(y \vert x)$.
However, the test data may come from a different distribution $p_\text{test}(x, y) = p_\text{test}(x) \cdot p(y \vert x)$.
For concreteness, we assume the conditional distribution $p(y \vert x)$ remains unchanged, but the marginal distribution of the input features $p_\text{test}(x)$ differs from $p_\text{train}(x)$; we note that our results do not depend on this particular definition of covariate shift. \citet{arjovsky2021out} provides a detailed discussion of covariate shift.

\section{Related work}

Methods to improve robustness to shift between train and test often explicitly make use of the test distribution in some fashion. For example, it is common to apply semi-supervised methods to the labelled training data augmented by the unlabelled test inputs \citep[e.g.,][]{daume2006domain, athiwaratkun2018there}, or to learn a shared feature transformation for both train and test \citep[e.g.,][]{daume2009frustratingly}. 
In a Bayesian setting, \citet{storkey2007mixture} and \citet{storkey2009training} propose such approaches for linear regression and Gaussian processes under covariate shift, assuming the data comes from multiple sources. 
Moreover, \citet{shimodaira2000improving} propose to re-weight the train data points according to their density in the test data distribution. 
\citet{sugiyama2006importance, sugiyama2007covariate} adapt this importance-weighting approach to the cross-validation setting. 

We focus on the setting of robustness to covariate shift without any access to the test distribution \citep[e.g.,][]{daume2006domain}. Bayesian methods are frequently applied in this setting, often motivated by the ability for a Bayesian model average to provide a principled representation of epistemic uncertainty: there are typically many consistent explanations for out of distribution points, leading to high uncertainty for these points.
Indeed, approximate inference approaches for Bayesian neural networks are showing good and increasingly better performance under covariate shift \citep[e.g.,][]{gal2016dropout, lakshminarayanan2017simple, ovadia2019can, maddox2019simple, wilson2020bayesian, dusenberry2020efficient, benton2021loss}.

Many works attempt to understand robustness to covariate shift. For example, for classical training \citet{neyshabur2020being} show that models relying on features that encode semantic structure in the data are more robust to covariate shift.
\citet{nagarajan2020understanding} also provide insights into how classically trained max-margin classifiers can fail under covariate shift due to a reliance on spurious correlations between class labels and input features. BNN robustness to adversarial attacks is a related area of study, but generally involves much smaller perturbations to the test covariates than covariate shift. \citet{carbone2020robustness} prove that BNNs are robust to gradient-based adversarial attacks in the large data, overparameterized limit, while \citet{wicker2021bayesian} present a framework for training BNNs with guaranteed robustness to adversarial examples.

In this work, we propose novel priors that improve robustness of BNNs under covariate shift.
In \citet{fortuin2021bayesian}, the authors explore a wide range of priors and, in particular, show that the distribution of the weights of SGD-trained networks is heavy-tailed such as Laplace and Student-$t$, and does not appear Gaussian.
Other heavy-tailed sparsity inducing priors have also been proposed in the literature \citep{carvalho2009handling, molchanov2017variational, kessler2019hierarchical, cui2020informative, izmailov2021bayesian, fortuin2021priors}.
Inspired by this work, we evaluate Laplace and Student-$t$ priors for BNNs, but find that they do not address the poor performance of BNNs under covariate shift. 

\citet{domingos2000bayesian}, \citet{minka2000bayesian}, \citet{masegosa2019learning} and \citet{morningstar2020pac} explore failure modes of Bayesian model averaging when the Bayesian model does not contain a reasonable solution in its hypothesis space, causing issues when the posterior contracts. This situation is orthogonal to the setting in our paper, where we know the Bayesian model does contain a reasonable solution in its hypothesis space, since the MAP estimate is robust to covariate shift. In our setting, robustness issues are caused by a \emph{lack} of posterior contraction.

In general, understanding and addressing covariate shift is a large area of study. For a comprehensive overview, see \citet{arjovsky2021out}. To our knowledge, no prior work has attempted to understand, further demonstrate, or remedy the poor robustness of Bayesian neural networks with high fidelity approximate inference recently discovered in \citet{izmailov2021bayesian}.

\section{Bayesian neural networks are not robust to covariate shift}
\label{sec:bnns_not_robust}

In this section, we evaluate Bayesian neural networks under different types of covariate shift.
Specifically, we focus on two types of covariate shift: test data corruption and domain shift.
In \autoref{appendix-sec:spurious_correlations}, we additionally evaluate BNNs in the presence of spurious correlations in the data. 

\textbf{Methods.}\quad
We evaluate BNNs against two deterministic baselines: 
a MAP solution approximated with stochastic gradient descent (SGD) with momentum \citep{robbins1951stochastic, polyak1964some} and
a deep ensemble of $10$ independently trained MAP solutions \citep{lakshminarayanan2017simple}.
For BNNs, we provide the results using a Gaussian prior and a more heavy-tailed Laplace prior following \citet{fortuin2021bayesian}.
\citet{izmailov2021bayesian} conjectured that \textit{cold posteriors} \citep{wenzel2020good} can improve the robustness of BNNs under covariate shift; to test this hypothesis, 
we provide results for BNNs with a Gaussian prior and cold posteriors at temperature $10^{-2}$.
For all BNN models, we run a single chain of HMC for $100$ iterations discarding the first $10$ iterations as burn-in, following \citet{izmailov2021bayesian}. We provide additional experimental details in \autoref{appendix-sec:exp_details}.

\textbf{Datasets and data augmentation.}\quad
We run all methods on the MNIST \citep{lecun2010mnist} and CIFAR-10 \citep{krizhevsky2014cifar} datasets.
Following \citet{izmailov2021bayesian} we do not use data augmentation with any of the methods, as it is not trivially compatible with
the Bayesian neural network framework \citep[e.g.,][]{izmailov2021bayesian, wenzel2020good}.

\textbf{Neural network architectures.}\quad
On both the CIFAR-10 and MNIST datasets we use a small convolutional network (CNN) inspired by LeNet-5 
\citep{lecun1998gradient}, with $2$ convolutional layers followed by $3$ fully-connected layers.
On MNIST we additionally consider a fully-connected neural network (MLP) with $2$ hidden layers of $256$ 
neurons each.
We note that high-fidelity posterior sampling with HMC is extremely computationally
intensive. Even on the small architectures that we consider, the experiments take multiple hours on
$8$ NVIDIA Tesla V-100 GPUs or 8-core TPU-V3 devices \citep{jouppi2020domain}.
See \citet{izmailov2021bayesian} for details on the computational requirements of full-batch HMC for
BNNs.

\begin{figure}
    \centering
    \includegraphics[width=0.95\textwidth]{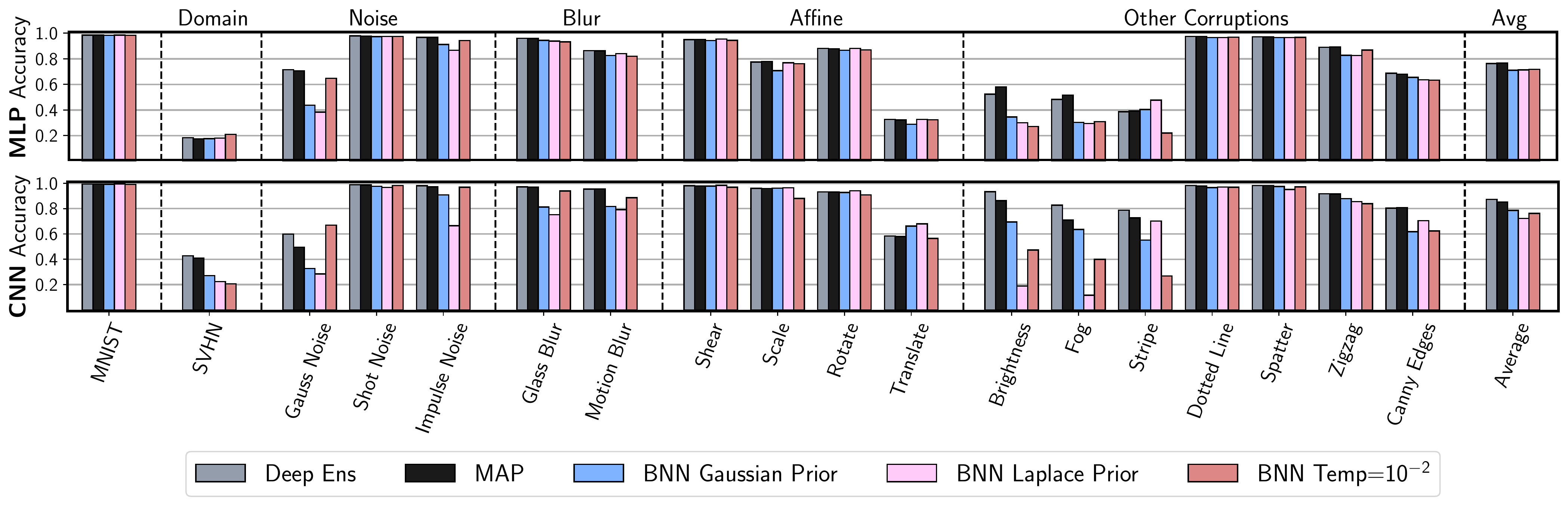}
    \caption{
    \textbf{Robustness on MNIST.}
    Accuracy for deep ensembles, MAP and Bayesian neural networks trained on MNIST under covariate shift.
    \textbf{Top}: Fully-connected network; \textbf{bottom}: Convolutional neural network.
    While on the original MNIST test set BNNs provide competitive performance, they underperform deep ensembles on most of the corruptions.
    With the CNN architecture, all BNN variants lose to MAP when evaluated on SVHN by almost $20\%$.
    }
     \label{fig:mnistc_accuracy}
\end{figure}

\subsection{Test data corruption}

We start by considering the scenario where the test data is corrupted by some type of noise.
In this case, there is no semantic distribution shift: the test data is collected in the same way as the train data, but then corrupted by a generic transformation.

In \autoref{fig:mnistc_accuracy}, we report the performance of the methods trained on MNIST and evaluated on the MNIST-corrupted (MNIST-C) test sets \citep{mu2019mnist} under various corruptions\footnote{In addition to the corruptions from MNIST-C, we consider Gaussian noise with standard deviation $3$, as this corruption was considered by \citet{izmailov2021bayesian}.}.
We report the results for a fully-connected network and a convolutional network.

With the CNN architecture, deep ensembles consistently outperform BNNs.
Moreover, even a single MAP solution significantly outperforms BNNs on many of the corruptions.
The results are especially striking on \textit{brightness} and \textit{fog} corruptions, where the BNN with Laplace prior shows accuracy at the level of random guessing, while the deep ensemble retains accuracy above $90\%$.
\textit{Gaussian noise} and \textit{impulse noise} corruptions also present significant challenges for BNNs.
The BNNs show the most competitive performance in-distribution and on the corruptions representing affine transformations:
\textit{shear}, \textit{scale}, \textit{rotate} and \textit{translate}.
The results are generally analogous for the MLP:
across the board the BNNs underperform deep ensembles, and often even a single MAP solution. 

While the cold posteriors provide an improvement on some of the \textit{noise} corruptions, they also hurt performance on \textit{Brightness} and \textit{Stripe}, and do not improve the performance significantly on average across all corruptions compared to a standard BNN with a Gaussian prior. We provide additional results on cold posterior performance in \autoref{appendix-tempering}.

Next, we consider the CIFAR-10-corrupted dataset (CIFAR-10-C) \citep{hendrycks2019benchmarking}.
CIFAR-10-C consists of $18$ transformations that are available at different levels of intensity (1 -- 5).  We report the results using corruption intensity $4$ (results for other intensities are in \autoref{appendix-priors}) for each of the transformations in \autoref{fig:cifarc_accuracy}.
Similarly to MNIST-C, BNNs outperform deep ensembles and MAP on in-distribution data, but underperform each over multiple corruptions. 
On CIFAR-10-C, BNNs are especially vulnerable to different types of \textit{noise} (\textit{Gaussian noise}, \textit{shot noise},
\textit{impulse noise}, \textit{speckle noise}).
For each of the \textit{noise} corruptions, BNNs underperform even the classical MAP solution.
The cold posteriors improve the performance on the \textit{noise} corruptions, but only provide a marginal improvement across the board. 

We note that \citet{izmailov2021bayesian} evaluated a ResNet-20 model on the same set of CIFAR-10-C corruptions.
While they use a much larger architecture, the qualitative results for both architectures are similar:
BNNs are the most vulnerable to \textit{noise} and \textit{blur} corruptions.
We thus expect that our paper's analysis is not specific to smaller architectures and will equally apply to deeper models.

\subsection{Domain shift}

\begin{figure}
    \centering
    \includegraphics[width=0.95\textwidth]{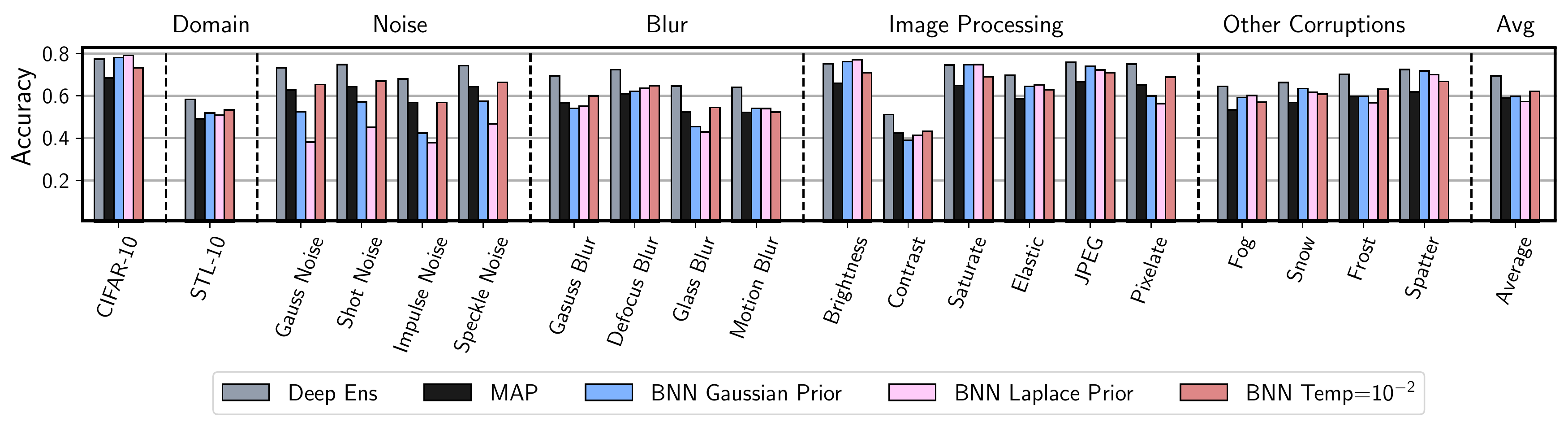}
    \caption{
    \textbf{Robustness on CIFAR-10.}
    Accuracy for deep ensembles, MAP and Bayesian neural networks using a CNN architecture trained on CIFAR-10 under covariate shift.
    For the corruptions from CIFAR-10-C, we report results for corruption intensity 4.
    While the BNNs with both Laplace and Gaussian priors outperform deep ensembles on the in-distribution accuracy,
    they underperform even a single MAP solution on most corruptions.
    }
     \label{fig:cifarc_accuracy}
\end{figure}

Next, we consider a different type of covariate shift where the test data and train data come from different, but
semantically related distributions.

First, we apply our CNN and MLP MNIST models to the SVHN test set \citep{netzer2011reading}.
The MNIST-to-SVHN domain shift task is a common benchmark for unsupervised domain adaptation:
both datasets contain images of digits, although MNIST contains hand-written digits while SVHN represents house numbers.
In order to apply our MNIST models to SVHN, we crop the SVHN images and convert them to grayscale. 
We report the results in \autoref{fig:mnistc_accuracy}.
While for MLPs all methods perform similarly, with the CNN architecture BNNs underperform deep ensembles and MAP by nearly $20\%$.

Next, we apply our CIFAR-10 CNN model to the STL-10 dataset \citep{coates2011analysis}.
Both datasets contain natural images with $9$ shared classes between the two datasets\footnote{CIFAR-10 has a class \textit{frog} and STL-10 has \textit{monkey}. The other nine classes coincide.}.
We report the accuracy of the CIFAR-10 models on these $9$ shared classes in STL-10 in \autoref{fig:cifarc_accuracy}.
While BNNs outperform the MAP solution, they still significantly underperform deep ensembles.

The results presented in this section highlight the generality and practical importance of the lack of robustness in BNNs:
despite showing strong performance in-distribution, BNNs underperform even a single MAP solution (classical training) over an extensive variety of covariate shifts.

\section{Understanding Bayesian neural networks under covariate shift}
\label{sec:linear_dependency}

Now that we have established that Bayesian neural networks are highly susceptible to many types of covariate shift, we seek to understand why this is the case. 
In this section, we identify the linear dependencies in the input features as one of the key issues undermining the robustness of BNNs. 
We emphasize that \textit{linear} dependencies in particular are \textit{not} simply chosen for the simplicity of analysis, and their key role follows from the structure of the fully-connected and convolutional layers. 

\subsection{Motivating example: dead pixels and fully-connected layers}
\label{subsec:dead_pixels}

To provide an intuition for the results presented in this section, we start with a simple but practically relevant motivating example.
Suppose we use a fully-connected Bayesian neural network on a dataset $D = \{(x_i, y_i)\}_{i=1}^n$,
with $m$ input features, where $x_i \in \mathbb R^m$.
We use the upper indices to denote the features $x^1, \ldots, x^m$ and the lower indices to denote the datapoints $x_i$.
Then, for each neuron $j$ in the first hidden layer of the network, the activation can be written as
$z_j = \phi(\sum_{i=1}^n x^i w^1_{ij} + b^1_j)$, where $w^1_{ij}$ is the weight of the first layer of the network corresponding to the input feature $i$ and hidden neuron $j$,
and $b^1_j$ is the corresponding bias. 
We show the following Lemma:

\begin{lemma}
\label{lemma:dead_pixel}
    Using the notation introduced above, suppose that the input feature $x^i_k$ is equal to zero for all the examples $x_k$ in the training dataset $D$.
    Suppose the prior distribution over the parameters $p(W)$ factorizes as $p(W) = p(w^1_{ij}) \cdot p(W \setminus w^1_{ij})$ for some 
    neuron $j$ in the first layer, where $W \setminus w^1_{ij}$ represents all the parameters $W$ of the network except $w^1_{ij}$.
    Then, the posterior distribution $p(W \vert D)$ will also factorize and the marginal posterior over the parameter $w^1_{ij}$ will coincide with the prior:
    \begin{equation}
        p(W \vert D) =
        p(W \setminus w^1_{ij} \vert D) \cdot p(w^1_{ij}).
    \end{equation}
    Consequently, the MAP solution will set the weight $w^1_{ij}$ to the value with maximum prior density.
\end{lemma}

Intuitively, Lemma \ref{lemma:dead_pixel} says that if the prior for some parameter in the network is independent of the other parameters, and the value of the parameter
does not affect the predictions of the model on any of the training data, then the posterior of this parameter will coincide with its prior. 
In particular, if one of the input features is always zero, then the corresponding weights will always be multiplied by zero, and will not affect the predictions of the network.
To prove Lemma \ref{lemma:dead_pixel}, we simply note that the posterior is proportional to the product of prior and likelihood, and both terms factorize with respect to $w^1_{ij}$.
We present a formal proof in \autoref{appendix-sec:theory}.

So, for any sample from the posterior, the weight $w^1_{ij}$ will be a random draw from the prior distribution.
Now suppose at test time we evaluate the model on data where the feature $x^i$ is no longer zero. Then for these new inputs, the model will be effectively multiplying the input feature 
$x^i$ by a random weight $w^1_{ij}$, leading to instability in predictions.
In \autoref{appendix-sec:theory}, we formally state and prove the following proposition:

\begin{customproposition}{1 (informal)}
\label{prop:dead_pixel}
    Suppose the assumptions of Lemma \ref{lemma:dead_pixel} hold.
    Assume also that the prior distribution $p(w^1_{ij})$ has maximum density at $0$ and that the network uses ReLU activations.
    Then for any test input $\bar x$, the expected prediction under Bayesian model averaging (\autoref{eq:bma}) will depend on the value of the feature $\bar x^i$, while the
    MAP solution will ignore this feature.
\end{customproposition}

For example, in the MNIST dataset there is a large number of \textit{dead pixels}: pixels near the boundaries of the image 
that have intensity zero for all the inputs. 
In practice, we often use independent zero-mean priors (e.g. Gaussian) for each parameter of the network.
So, according to Lemma \ref{lemma:dead_pixel}, the posterior over all the weights in the first layer of the network corresponding to the
dead pixels will coincide with the prior.
If at test time the data is corrupted by e.g., Gaussian noise, the dead pixels will receive non-zero intensities, leading to a significant
degradation in the performance of the Bayesian model average compared to the MAP solution.
 
While the situation where input features are constant and equal to zero may be rare and easily addressed, the results presented in this section
can be generalized to \emph{any} linear dependence in the data. We will now present our results in the most general form.

\subsection{General linear dependencies and fully-connected layers} 
\label{subsec:theory}

We now present our general results for fully-connected Bayesian neural networks when the features are linearly dependent.
Intuitively, if there exists a direction in the input space such that all of the training data points have a constant projection on this direction (i.e. the data lies in a hyper-plane), then posterior coincides with the prior in this direction.
Hence, the BMA predictions are highly susceptible to perturbations that move the test inputs in a direction orthogonal to the hyper-plane.
The MAP solution on the other hand is completely robust to such perturbations.
In \autoref{appendix-sec:theory} we prove the following proposition.

\begin{proposition}
\label{prop:fc_general}
    Suppose that the prior over the weights $w_{ij}^1$ and biases $b_{j}^1$ in the first layer is an i.i.d. Gaussian distribution $\mathcal N(0, \alpha^2)$,
    independent of the other parameters in the model. 
    Suppose all the inputs $x_1 \ldots x_n$ in the training dataset $D$ lie in an affine subspace of the input space:
    $\sum_{j=1}^m x_i^j c_j = c_0$ for all $i = 1, \ldots, n$ and some constants $c$ such that $\sum_{i=0}^m c_i^2 = 1$.
    Then,
    \begin{enumerate}
    \item[(1)] For any neuron $j$ in the first hidden layer, the posterior distribution of random variable $w^{c}_j = \sum_{i=1}^{m} c_i w^1_{ij} - c_0 b_j^1$ (the projection of parameter vector $(w_{1j}^1, \ldots, w_{mj}^1, b_j^1)$ on direction $(c_1, \ldots, c_m, -c_0)$) will coincide with the prior $\mathcal N(0, \alpha^2)$. 
    \item[(2)] The MAP solution will set $w^{c}_j$ to zero.
    \item[(3)] (Informal) Assuming the network uses ReLU activations, at test time, the BMA prediction will be susceptible to the inputs $\bar x$ that lie outside of the subspace, i.e. the predictive mean will
    depend on $\sum_{j=1}^m \bar x^j c_j - c_0$. 
    The MAP prediction will not depend on this difference.
    \end{enumerate}
\end{proposition}

\begin{figure}
\centering
    \begin{tabular}{cc}
        \hspace{-0.3cm}\includegraphics[height=0.195\textwidth]{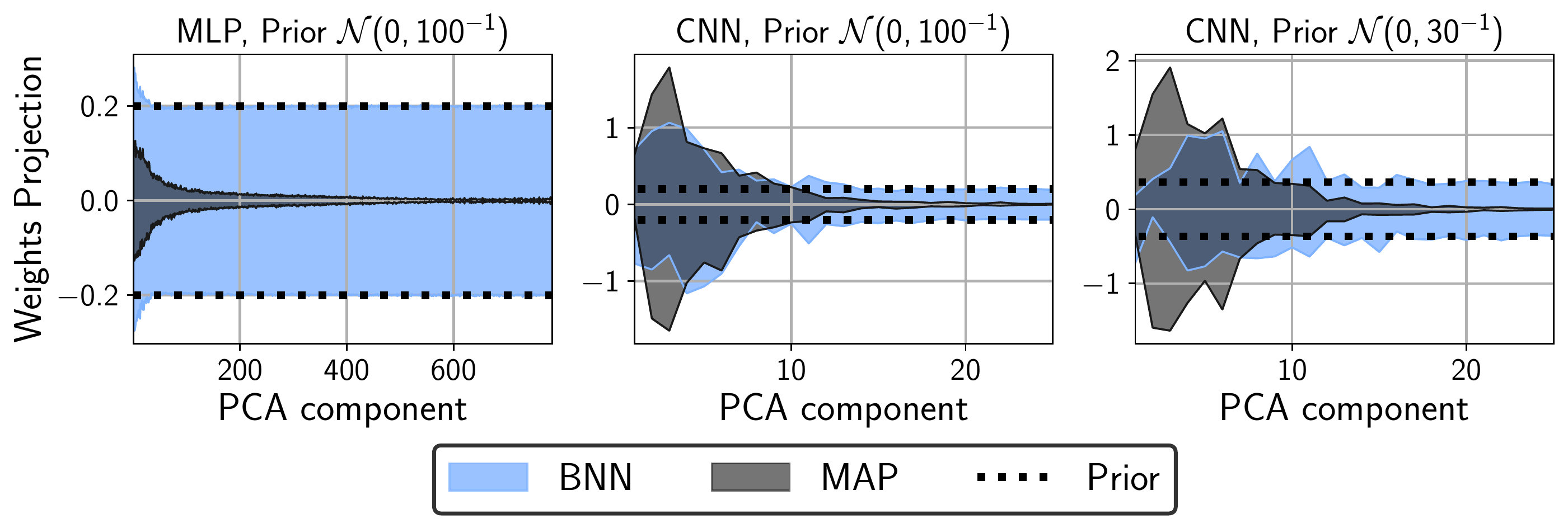}
        &
        \includegraphics[height=0.195\textwidth]{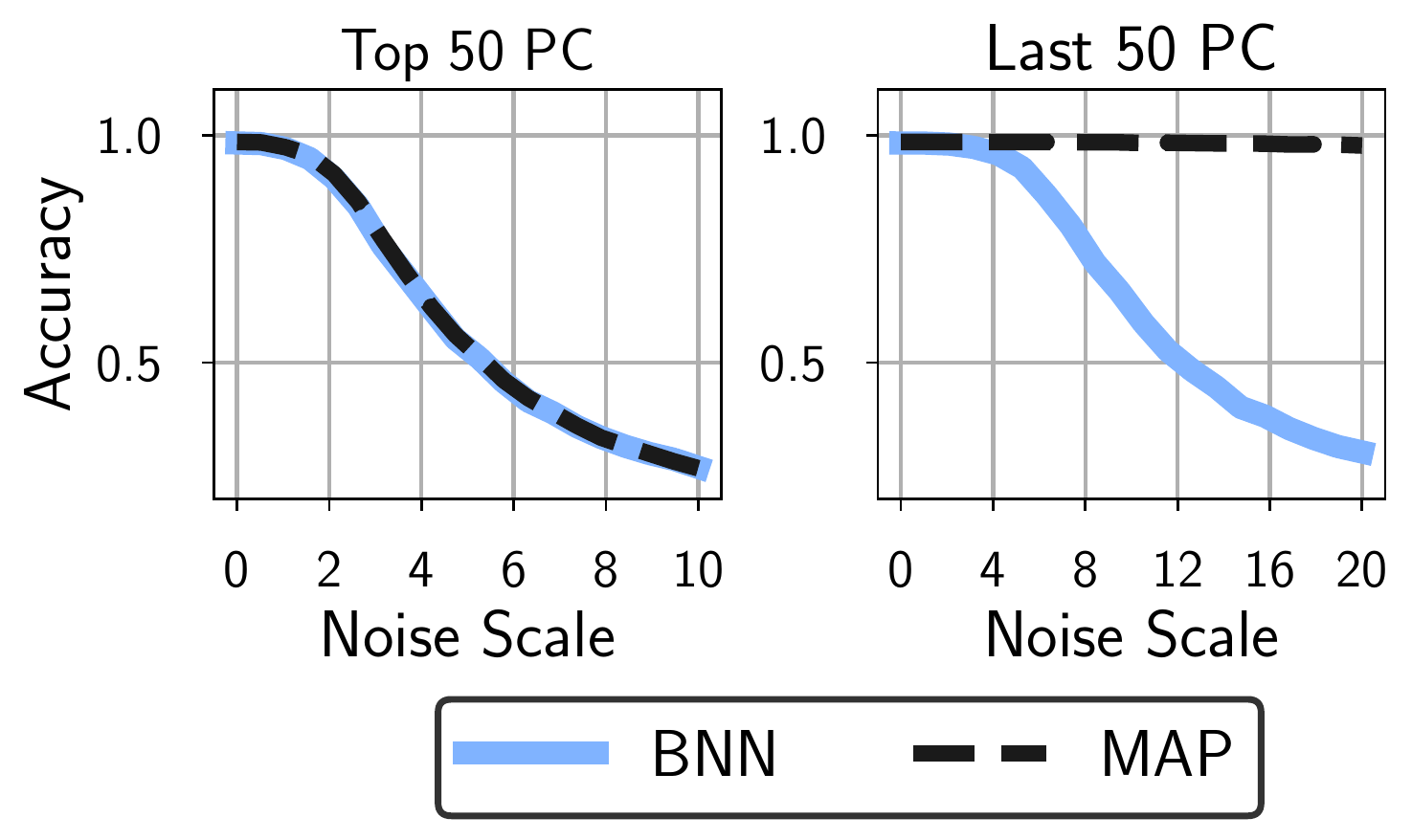}
      \\
        \hspace{-0.3cm}{\small (a) Weight projections} & 
        \hspace{-0.cm}{\small (b) Robustness along PCs} 
    \end{tabular}
\caption{
\textbf{\textbf{Bayesian inference samples weights along low-variance principal components from the prior, while MAP sets these weights to zero.}}
\textbf{(a)}: The distribution (mean $\pm$ 2 std) of projections of the weights of the first layer on the directions corresponding to the PCA components of the data for BNN samples and MAP solution using MLP and CNN architectures with different prior scales.
In each case, MAP sets the weights along low-variance components to zero, while BNN samples them from the prior.
\textbf{(b)}: Accuracy of BNN and MAP solutions on the MNIST test set with Gaussian noise applied along the 50 highest and 50 lowest variance PCA components of the train data (left and right respectively).  
MAP is very robust to noise along low-variance PCA directions, while BMA is not; the two methods are similarly robust along the 
highest-variance PCA components.
}
\label{fig:empirical_analysis}
\end{figure}

\textbf{Empirical support.}\quad
To test Proposition \ref{prop:fc_general}, we examine the performance of a fully-connected BNN on MNIST. The MNIST training dataset is not full rank, meaning that it has linearly dependent features. For a fully-connected BNN, Proposition \ref{prop:fc_general} predicts that the posterior distribution of the first layer weights projected onto directions corresponding to these linearly dependent features will coincide with the prior. In \autoref{fig:empirical_analysis}(a) we test this hypothesis by projecting first layer weights onto the principal components of the data.
As expected, the distribution of the projections on low-variance PCA components (directions that are constant or nearly constant in the data) almost exactly coincides with the prior.
The MAP solution, on the other hand, sets the weights along these PCA components close to zero, confirming conclusion (2) of the proposition.
Finally, in \autoref{fig:empirical_analysis}(b) we visualize the performance of the BMA and MAP solution as we apply noise along high-variance and low-variance directions in the data.
As predicted by conclusion (3) of Proposition \ref{prop:fc_general}, the MAP solution is very robust to noise along the low-variance directions, while BMA is not.

\subsection{Linear dependencies and convolutional layers}

Finally, we can extend Proposition \ref{prop:fc_general} to convolutional layers.
Unlike fully-connected layers, convolutional layers include weight sharing such that no individual weight corresponds to a specific pixel in the input images.
For example, dead pixels will not necessarily present an issue for convolutional layers, unlike what is described in \autoref{subsec:dead_pixels}.
However, convolutional layers are still susceptible to a special type of linear dependence.
Intuitively, we can think of the outputs of the first convolutional layer on all the input images as the outputs of a fully-connected layer applied to all the $K \times K$ patches 
of the input image.
Therefore the reasoning in Proposition \ref{prop:fc_general} applies to the convolutional layers, with the difference that the linear dependencies in the $K \times K$ patches cause the instability rather than dependencies in the full feature space.
In \autoref{appendix-sec:theory} we prove the following proposition.

\begin{proposition}
\label{prop:conv_general}
    Suppose that the prior over the parameters of the convolutional filters and biases in the first layer is an i.i.d. Gaussian distribution $\mathcal N(0, \alpha^2)$,
    independent of the other parameters in the model. 
    Suppose that the convolutional filters in the first layer are of size $K \times K \times C$, where $C$ is the number of input channels.
    Then, consider the set $\hat D$ of size $N$ of all the patches of size $K \times K \times C$ extracted from the training images in $D$ after applying the same
    padding as in the first convolutional layer.
    Suppose all the patches $z_1 \ldots z_N$ in the dataset $\hat D$ lie in an affine subspace of the space $\mathbb{R}^{K \times K \times C}$:
    $\sum_{c=1}^C \sum_{a=1}^K \sum_{b=1}^K z_i^{a, b} \gamma_{c, a, b} = \gamma_0$ for all $i = 1, \ldots, N$ and some constants $c_i$ such that $\sum_{c=1}^C \sum_{a=1}^K \sum_{b=1}^K \gamma_{c, a, b}^2 + \gamma_0^2 = 1$.
    Then, we can prove results analogous to (1)-(3) in Proposition \ref{prop:fc_general} (see the \autoref{appendix-sec:theory} for the details).
\end{proposition}

\textbf{Empirical support.}\quad
In \autoref{fig:empirical_analysis}(a), we visualize the projections of the weights in the first layer of the CNN architecture on the PCA components of the $K \times K$ patches extracted from MNIST.
Analogously to fully-connected networks, the projections of the MAP weights are close to zero for low-variance components,
while the projections of the BNN samples follow the prior.

\subsection{What corruptions will hurt performance?}
Based on Propositions \ref{prop:fc_general}, \ref{prop:conv_general}, we expect that the corruptions that are the most likely to break linear dependence
structure in the data will hurt the performance the most. 
In \autoref{appendix-sec:corruptions_vs_pca} we argue that \textit{noise} corruptions are likely to break linear dependence, while 
the \textit{affine} corruptions are more likely to preserve it, agreeing with our observations in \autoref{sec:bnns_not_robust}.

\subsection{Why do some approximate Bayesian inference methods work well under covariate shift?}

Unlike BNNs with HMC inference, some approximate inference methods such as SWAG \citep{maddox2019simple}, MC dropout \citep{gal2016dropout}, deep ensembles \citep{lakshminarayanan2017simple} and mean field VI \citep{blundell2015weight} provide strong performance under covariate shift \citep{ovadia2019can, izmailov2021bayesian}. 
For deep ensembles, we can easily understand why: a deep ensemble represents an average of approximate MAP solutions, and we have seen in conclusion (3) of Proposition \autoref{prop:fc_general} that MAP is robust to covariate shift in the scenario introduced in \autoref{subsec:theory}. 
Similarly, other methods are closely connected to MAP via characterizing the posterior using MAP optimization iterates (SWAG), or modifying the training procedure for the MAP solution (MC Dropout). 
We provide further details, including theoretical and empirical analysis of variational inference under covariate shift, in \autoref{appendix-sec:approximate_inference}.

\section{Towards more robust Bayesian model averaging under covariate shift}
\label{sec:solution}

In this section, we propose a simple new prior inspired by our theoretical analysis.
In \autoref{sec:linear_dependency} we showed that linear dependencies in the input features cause the posterior to coincide
with the prior along the corresponding directions in the parameter space.
In order to address this issue, we explicitly design the prior for the first layer of the network so that the variance is low along these directions.

\subsection{Data empirical covariance prior}

Let us consider the empirical covariance matrix of the inputs $x_i$.
Assuming the input features are all preprocessed to be zero-mean $\sum_{i=1}^n x_i = 0$, we have
$\Sigma = \frac{1}{n-1}\sum_{i=1}^n x_i x_i^T$.
For fully-connected networks, we propose to use the \textit{EmpCov} prior $p(w^1) = \mathcal N(0, \alpha \Sigma + \epsilon I)$ on the weights $w^1$ of the first layer of the network, where $\epsilon$ is a small positive constant ensuring that the covariance matrix is positive definite.
The parameter $\alpha > 0$ determines the scale of the prior.

Suppose there is a linear dependence in the input features of the data: $x_i^T p = c$ for some direction $p$ and constant $c$.
Then $p$ will be an eigenvector of the empirical covariance matrix with the corresponding eigenvalue equal to $0$ : $\Sigma p = \frac{1}{n-1}\sum_{i=1}^n x_i x_i^T p = \frac{c}{n-1}\sum_{i=1}^n x_i = 0$.
Hence the prior over $w^1$ will have a variance of $\epsilon$ along the direction $p$.

More generally, the \textit{EmpCov} prior is aligned with the principal components of the data, which are the eigenvectors of the matrix $\Sigma$.
The prior variance along each principal component $p_i$ is equal to $\alpha \sigma_i^2 + \epsilon$ where $\sigma_i^2$ is its corresponding explained variance.
In \autoref{sec:general_pca_priors}, we discuss a more general family of priors aligned with the principal components of the data.

\textbf{Generalization to convolutions.}\quad
We can generalize the \textit{EmpCov} prior to convolutions by replacing the empirical covariance of the data with the empirical covariance of the patches that interact with the convolutional filter, denoted by $\hat D$ in Proposition \ref{prop:conv_general}. 

\textbf{Is EmpCov a valid prior?}\quad
\emph{EmpCov} constructs a valid prior by evaluating the empirical covariance matrix of the inputs. This prior does not depend on the train data labels $y_i$, unlike the approach known as \textit{Empirical Bayes} \citep[see e.g.][section 3.5]{bishop06}, which is commonly used to specify hyperparameters in Gaussian process and neural network priors \citep{rasmussen06, mackay1995probable}.

\subsection{Experiments}
\label{sec:empcov_prior_experiments}

\begin{figure}
\centering
\includegraphics[width=0.95\textwidth]{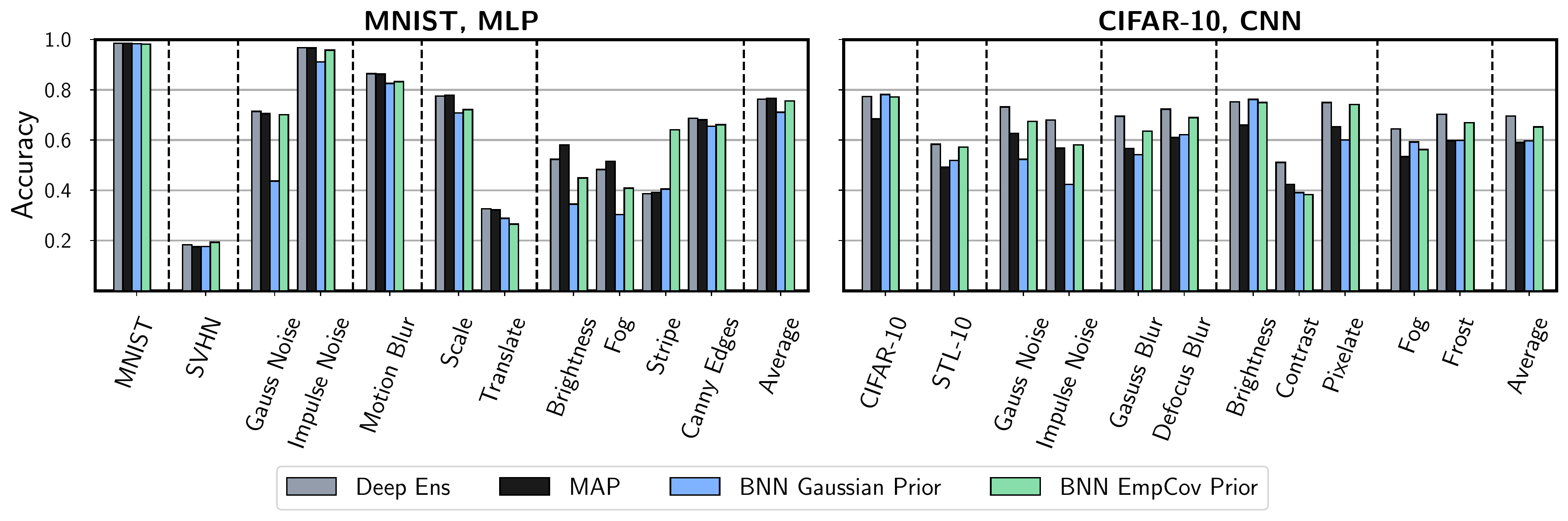}
\caption{
\textbf{EmpCov prior improves robustness.}
Test accuracy under covariate shift for deep ensembles, MAP optimization with SGD, and BNN with Gaussian and \textit{EmpCov} priors.
\textbf{Left}: MLP architecture trained on MNIST. 
\textbf{Right}: CNN architecture trained on CIFAR-10.
The \textit{EmpCov} prior provides consistent improvement over the standard Gaussian prior.
The improvement is particularly noticeable on the \textit{noise} corruptions and domain shift experiments (SVHN, STL-10).
}
\label{fig:empcov}
\end{figure}

In \autoref{fig:empcov} we report the performance of the BNNs using the \textit{EmpCov} prior.
In each case, we apply the \textit{EmpCov} prior to the first layer, and a Gaussian prior to all other layers.
For more details, please see \autoref{appendix-sec:exp_details}.
On both MLP on MNIST and CNN on CIFAR-10, the \textit{EmpCov} prior significantly improves performance across the board.
In both cases, the BNN with \textit{EmpCov} prior shows competitive performance with deep ensembles, especially in the domain shift experiments.
\textit{EmpCov} is also particularly useful on the \textit{noise} corruptions.

In \autoref{sec:nonzero_mean_noise}, we provide a detailed analysis of the performance of the CNN architecture on MNIST.
Surprisingly, we found that using the \textit{EmpCov} prior by itself does not provide a large improvement in this case.
In \autoref{sec:nonzero_mean_noise}, we identify an issue specific to this particular setting, and propose another targeted prior that substantially improves performance.

\section{Discussion}
\label{sec:understanding_extensions}
\vspace{-2mm}

We consider the generality of our results, additional perspectives, and future directions.

\textbf{Is the poor generalization of BNNs under covariate shift surprising?}\quad
The results presented in this paper are of crucial practical importance, relevant to the applicability of Bayesian neural networks in virtually every real-world setting since the train distribution is often not exactly the same as the test distribution. Ultimately, whether or not a result is \emph{surprising} is subjective, but there are many reasons to find the dramatic performance degradation of Bayesian neural networks under shift surprising, which we outline in 
\autoref{appendix-surprising}.

\textbf{Why focus on linear dependencies?}\quad
This focus is dictated by the structure of both fully-connected and convolutional neural networks, which apply activation functions to linear combinations of features.
Our results can be directly extended to other models, where different types of dependencies between input features would lead to the same lack of robustness to covariate shift.
For example, in \autoref{sec:bayesian_nalu} we derive analogous results to \autoref{subsec:theory} for multiplicative neural network architectures \citep{trask2018neural},
where a different form of dependence between features causes poor robustness.

\textbf{Intermediate layers.}\quad
While our analysis in \autoref{sec:linear_dependency} is focused on the linear dependencies in the input features, similar conclusions can be made about intermediate layers of the network. For example, weights connected to \emph{dead neurons}, which output zero, do not affect predictions of the model. We provide more details in \autoref{sec:dead_neurons}. 

\textbf{Linear Bayesian models.}\quad
In models that are linear in parameters, such as Bayesian linear regression and Gaussian process regression, we are typically saved from the perils of BMA under covariate shift discussed in Section~\ref{sec:linear_dependency}, because the MAP and the predictive mean under BMA coincide. 
We provide further details in \autoref{sec:app_blr}.

\textbf{BNNs in low-data regime.}\quad
While we focus on covariate shift, our results also suggest that BNNs may generalize poorly when the training dataset is extremely small.
Indeed, in low-data regime we may observe linear combinations of the features that are constant in the training data, but not on test.
In \autoref{appendix-lowdata} we show empirically that BNNs can underperform MAP in low-data regime.

\textbf{An optimization perspective on the covariate shift problem.}\quad
In Section~\ref{subsec:dead_pixels}, we argued that SGD is more robust to covariate shift than HMC because the regularizer pushes the weights that correspond to dead pixels towards zero. This effect is mainly obtained through explicit regularization, without which these weights will remain at their initial values. 
We study the effect of initialization and explicit regularization on the performance of SGD under covariate shift in Appendix \ref{sec:init-reg-effect}. We also 
show in Appendix \ref{sec:other-optimizers} that other stochastic optimizers such as Adam~\citep{kingma2014adam} and Adadelta~\citep{zeiler2012adadelta} behave similarly to SGD under covariate shift. Finally, we discuss the effect of test data corruptions on the loss landscape in Appendix \ref{sec:loss-surface} and argue that the relative sharpness of low density posterior samples makes these solutions, which are included in a BMA but not MAP, more vulnerable to covariate shift.

\textbf{Limitations.}\quad Due to the intense computational requirements of HMC, our experiments are limited to smaller models and datasets. Our analysis is focused on issues arising in models that are non-linear in their parameters. Moreover, while our proposed priors help improve robustness, they do not entirely resolve the issue. For example, in the \emph{Contrast} dataset of Figure~\ref{fig:empcov} (right panel), the BMA is still underperforming MAP. 

\textbf{Conclusion.}\quad
Our work has demonstrated, both empirically and theoretically, how linear dependencies in the training data cause Bayesian neural networks to generalize poorly under covariate shift --- explaining the important and unexpected findings in \citet{izmailov2021bayesian}. The scope of this research is exceptionally broad, relevant to the safe deployment of Bayesian methods in virtually any real-world setting.
While the two priors we introduce achieve some improvement in BNN performance under covariate shift, we are only beginning to explore possible remedies. Our work is intended as a step towards understanding the true properties of Bayesian neural networks, and improving the robustness of Bayesian model averaging under covariate shift. 

\section*{Acknowledgements}
We thank Martin Arjovsky, Behnam Neyshabur, Vaishnavh Nagarajan, Marc Finzi, Polina Kirichenko, Greg Benton and Nate Gruver for helpful discussions. 
This  research  is  supported with Cloud TPUs from Google's TPU Research Cloud (TRC), and by  an  Amazon  Research  Award, NSF  I-DISRE  193471,  NIH R01DA048764-01A1,  NSF IIS-1910266,  and NSF 1922658 NRT-HDR: FUTURE Foundations,  Translation,  and Responsibility for Data Science.

\bibliography{references}
\bibliographystyle{plainnat}

\newpage

\appendix

\section*{Appendix Outline}
This appendix is organized as follows. 
\begin{itemize}
\item In \autoref{appendix-sec:exp_details}, we provide details on hyper-parameters, datasets and architectures used in our experiments.
\item In \autoref{appendix-surprising}, we discuss whether the poor generalization of BNNs under covariate shift is surprising.
\item In \autoref{appendix-priors}, we examine BNN performance under covariate shift for a variety of different standard priors with different hyper-parameter settings. 
\item In \autoref{appendix-sec:spurious_correlations}, we study the effect of spurious correlations on BMA performance using the shift-MNIST dataset. 
\item In \autoref{appendix-lowdata}, we show that the same issues that hurt BNN generalization under covariate shift can cause poor performance in low-data regime.
\item In \autoref{appendix-convergence}, we explore the convergence of the BNN performance as a function of the number of HMC samples we produce.
\item In \autoref{appendix-tempering}, we study how temperature scaling impacts BMA performance under covariate shift. 
\item In \autoref{appendix-sec:theory}, we provide proofs of our propositions from \autoref{sec:linear_dependency}. 
\item In \autoref{appendix-sec:corruptions_vs_pca}, we visualize how different corruptions introduce noise along different principal components of the data, and relate this to BMA performance on these corruptions. 
\item In \autoref{appendix-sec:approximate_inference}, we explain why approximate inference methods SWAG and MC Dropout do not suffer the same performance degradation under covariate shift as HMC. 
\item In \autoref{sec:general_pca_priors}, we analyze a more general family of priors that includes the \textit{EmpCov} prior from \autoref{sec:solution}. 
\item In \autoref{sec:nonzero_mean_noise}, we introduce the sum filter prior for improving BNN robustness to non-zero-mean noise. 
\item In \autoref{sec:bayesian_nalu}, we provide an example of a model architecture where the BMA will be impacted by nonlinear dependencies in the training data. 
\item In \autoref{sec:dead_neurons}, we examine how BNNs can be impacted by linear dependencies beyond the first layer using the example of dead neurons. 
\item In \autoref{sec:app_blr} we prove that linear dependencies do not hurt BMAs of linear models under covariate shift. 
\item In \autoref{sec:optimization-perspective}, we examine covariate shift from an optimization perspective. 
\item Lastly, in \autoref{sec:app_licenses}, we provide details on licensing.
\end{itemize}

\section{Hyper-parameters and details of experiments}
\label{appendix-sec:exp_details}

\subsection{Prior definitions}
\label{sec:app_prior_defs}

Here we define the prior families used in the main text and the appendix, and the corresponding hyper-parameters.

\textbf{Gaussian priors.}
\quad
We consider iid Gaussian priors of the form $\mathcal N(0, \alpha^2 I)$, where $\alpha^2$ is the prior variance.
Gaussian priors are the default choice in Bayesian neural networks \citep[e.g.][]{fortuin2021bayesian, izmailov2021bayesian, wilson2020bayesian}.

\textbf{Laplace priors.}
\quad
We consider priors of the form $\text{Laplace}(\alpha): \frac 1 {2\alpha} \exp(-\|x\|_1 / \alpha)$, where $\|\cdot\|_1$ is the $\ell_1$-norm.

\textbf{Student-t priors.}
\quad
In \autoref{appendix-priors}, we consider iid Student-t priors of the form \\
$\text{Student-}t(\nu, \alpha^2): \frac{\Gamma (\frac{\nu+1}{2})}{\Gamma(\frac{\nu}{2})\sqrt{\nu \pi}}(1+\frac{w^2}{\nu \alpha^2})^{-\frac{\nu+1}{2}}$, where $\nu$ represents the degrees of freedom and $\alpha^2$ is the prior variance. 

\textbf{Exp-norm priors.}
\quad
In \autoref{appendix-priors}, we consider the prior family of the form\\ $\text{ExpNorm}(p, \alpha^2): \exp(-\|w\|^p / {2 \alpha^2})$.
Notice that for $p=2$, we get the Gaussian prior family.
By varying $p$ we can construct more heavy-tailed ($p < 2$) or less heavy-tailed ($p>2$) priors.

\subsection{Hyper-parameters and details}

\textbf{HMC hyper-parameters.}\quad
The hyper-parameters for HMC are the step size, trajectory length and any hyper-parameters of the prior. 
Following \citet{izmailov2021bayesian}, we set the trajectory length $\tau = \frac{\pi \sigma_{prior}}{2}$ where $\sigma_{prior}$ is the standard deviation of the prior. 
We choose the step size to ensure that the accept rates are high; for most of our MLP runs we do $10^4$ leapfrog steps per sample, while for CNN we do $5 \cdot 10^3$ leapfrog steps per sample.
For each experiment, we run a single HMC chain for 100 iterations discarding the first 10 iterations as burn-in;
in \autoref{appendix-convergence} we show that 100 samples are typically sufficient for convergence of the predictive performance.

\textbf{Data Splits.}\quad
For all CIFAR-10 and MNIST experiments, we use the standard data splits: 50000 training samples for CIFAR-10, 60000 training samples for MNIST, and 10000 test samples for both. For all data corruption experiments, we evaluate on the corrupted 10000 test samples. For domain shift experiments, we evaluate on 26032 SVHN test samples for MNIST to SVHN and 7200 STL-10 test samples for CIFAR-10 to STL-10.
In all cases, we normalize the inputs using train data statistics, and do not use any data augmentation.

\textbf{Neural network architectures.}\quad
Due to computational constraints, we use smaller neural network architectures for our experiments. All architectures use ReLU activations.
For MLP experiments, we use a fully-connected network with 2 hidden layers of 256 neurons each.
For CNN experiments, we use a network with 2 convolutional layers followed by 3 fully-connected layers. 
Both convolutional layers have $5 \times 5$ filters, a stride of 1, and use $2 \times 2$ average pooling with stride 2. The first layer has 6 filters and uses padding, while the second layer has 16 filters and does not use padding. The fully connected layers have 400, 120, and 84 hidden units.

\textbf{MAP and deep ensemble hyper-parameters.}\quad
We use the SGD optimizer with momentum $0.9$, cosine learning rate schedule and weight decay $100$ to approximate the MAP solution.
In \autoref{sec:optimization-perspective} we study the effect of using other optimizers and weight decay values.
On MNIST, we run SGD for $100$ epochs, and on CIFAR we run for $300$ epochs.
For the deep ensemble baselines, we train $10$ MAP models independently and ensemble their predictions. 

\textbf{Prior hyper-parameters.}\quad
To select prior hyper-parameters we perform a grid search, and report results for the optimal hyperparameters in order to compare the best versions of different models and priors. 
We report the prior hyper-parameters used in our main evaluation in \autoref{table:prior_hypers}.
In \autoref{appendix-priors} we provide detailed results for various priors with different hyper-parameter choices. 

\renewcommand{\arraystretch}{1.5}
\begin{table}[h]
  \caption{Prior hyper-parameters}
  \label{table:prior_hypers}
  \centering
  \begin{tabular}{lccc}
    \toprule
    Hyper-parameter     & MNIST MLP & MNIST CNN & CIFAR-10 CNN \\
    \midrule
    BNN, Gaussian prior; $\alpha^2$ & $\frac 1 {100}$ & $\frac 1 {100}$ & $\frac 1 {100}$ \\
    BNN, Laplace prior; $\alpha$ &  $\sqrt{\frac 1 6}$ & $\sqrt{\frac 1 6}$ & $\sqrt{\frac 1 {200}}$ \\
    \bottomrule
  \end{tabular}
\end{table}

\textbf{Tempering hyper-parameters.}\quad
For the tempering experiments, we use a Gaussian prior with variance $\alpha^2 = \frac 1 {100}$ on MNIST and $\alpha^2 = \frac 1 3$ on CIFAR-10.
We set the posterior temperature to $10^{-2}$.
We provide additional results for other prior variance and temperature combinations in \autoref{appendix-tempering}.

\textbf{Compute.}\quad
We ran all the MNIST experiments on TPU-V3-8 devices, and all CIFAR experiments on 8 NVIDIA Tesla-V100 devices.
A single HMC chain with $100$ iterations on these devices takes roughly 1.5 hours for MNIST MLP, 2 hours for CIFAR CNN and 3 hours for MNIST CNN.
As a rough upper-bound, we ran on the order of $100$ different HMC chains, each taking $2$ hours on average, resulting in $200$ hours on our devices, or
roughly $1600$ GPU-hours (where we equate 1 hour on TPU-V3-8 to 8 GPU-hours).

\section{Should we find the lack of BMA robustness surprising?}
\label{appendix-surprising}

Bayesian neural networks are sometimes presented as a way of improving \textit{just} the uncertainties, often at the cost of degradation in accuracy.
Consequently, one might assume that the poor performance of BNNs under covariate shift is not surprising, and we should use BNN uncertainty estimates solely to \textit{detect} OOD, without attempting to make predictions, even for images that are still clearly recognizable.

In recent years, however, Bayesian deep learning methods \citep[e.g.,][]{maddox2019simple, dusenberry2020efficient, daxberger2020expressive},
as well as high-fidelity approximate inference with HMC \citep{izmailov2021bayesian}, achieve improved uncertainty \textit{and} accuracy compared to standard MAP training with SGD.
In this light, we believe there are many reasons to find the significant performance degradation under shift surprising:
\begin{itemize}
    \item The BNNs are often providing significantly better accuracy on in-distribution points. For example, HMC BNNs achieve a 5\% improvement over MAP on CIFAR-10, but 25\% worse accuracy on the pixelate corruption, when the images are still clearly recognizable (see \autoref{fig:intro_figure}). To go from clearly better to profoundly worse would not typically be expected of any method on these shifts.
    \item In fact, recent work \citep[e.g.][]{miller2021accuracy} shows that there is typically a strong correlation between in-distribution and OOD generalization accuracy on related tasks, which is the opposite of what we observe in this work. 
    \item Many approximate Bayesian inference procedures do improve accuracy over MAP on shift problems \citep{ovadia2019can, wilson2020bayesian,dusenberry2020efficient}, and newer inference procedures appear to be further improving on these results. For example, MultiSWAG \citep{wilson2020bayesian} is significantly more accurate than MAP under shift. The fact that these methods are more Bayesian than MAP, and improve upon MAP in these settings, makes it particularly surprising that a high-fidelity BMA would be so much worse than MAP. This is a nuanced point — how is it that methods getting closer in some ways to the Bayesian ideal are improving on shift, when a still higher-fidelity representation of the Bayesian ideal is poor on shift? — we discuss this point in \autoref{appendix-sec:approximate_inference}. 
    \item Recent results highlight that there need not be a tension between OOD detection and OOD generalization accuracy: indeed deep ensembles provide much better performance than MAP on both \citep{ovadia2019can}. 
    \item Bayesian methods are closely associated with trying to provide a good representation of uncertainty, and a good representation of uncertainty should not say “I have little idea” when a point is only slightly out of distribution, but still clearly recognizable, e.g., through noise corruption or mild domain shift. 
    \item In \autoref{fig:detailed_results} we report the log-likelihood and ECE metrics which evaluate the quality of uncertainty estimates for deep ensembles, MAP and BNNs. The log-likelihood and ECE of standard BNNs are better than the corresponding values for the MAP solution on average, but they are much worse than the corresponding numbers for deep ensembles for high degrees of corruption. Furthermore, for some corruptions (\textit{impulse noise}, \textit{pixelate}) BNNs lose to MAP on both log-likelihood and ECE at corruption intensity 5. Also for larger ResNet-20 architecture on CIFAR-10-C, \citet{izmailov2021bayesian} reported that the log-likelihoods of BNNs are on average slightly worse than for MAP solution at corruption intensity 5.
\end{itemize}

\section{Additional results on BNN robustness}
\label{appendix-priors}

\subsection{Error-bars and additional metrics}

\begin{figure}
    \centering
    \begin{subfigure}[t]{\textwidth}
        \includegraphics[width=0.99\textwidth]{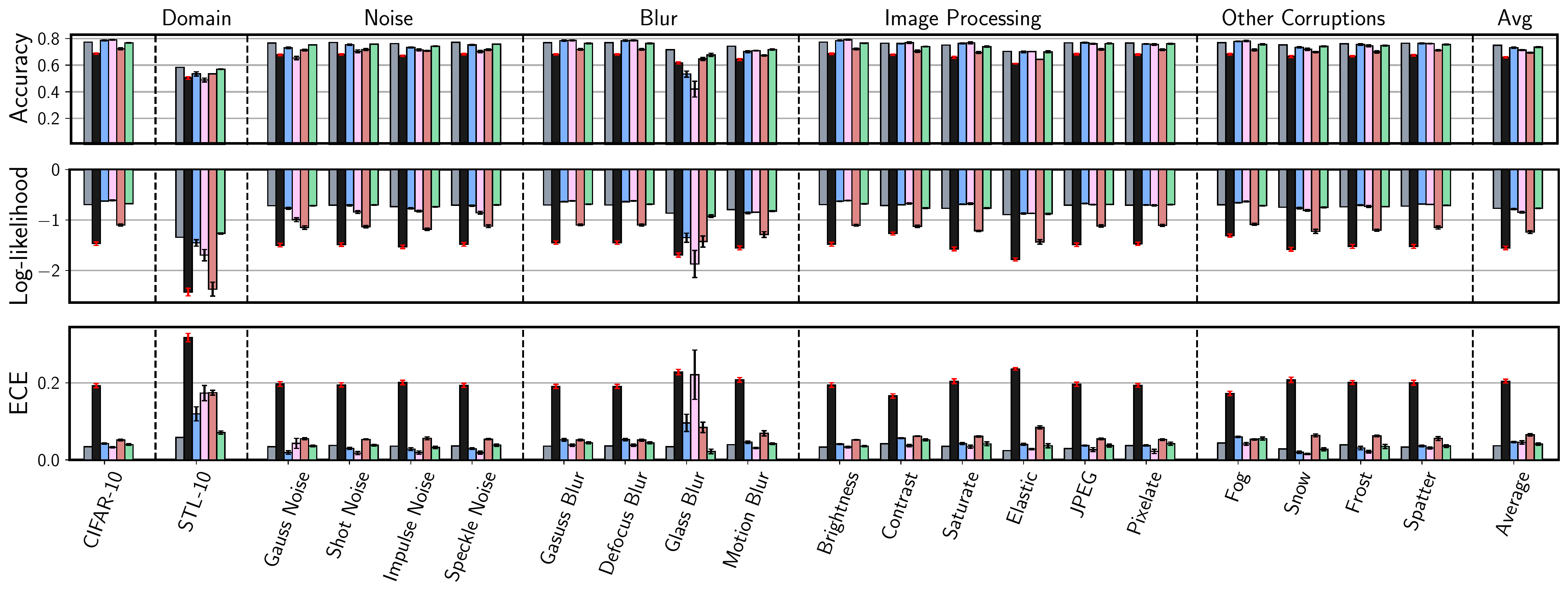}
        
        \caption{Corruption Intensity 1}
    \end{subfigure}
    \begin{subfigure}[t]{\textwidth}
        \includegraphics[width=0.99\textwidth]{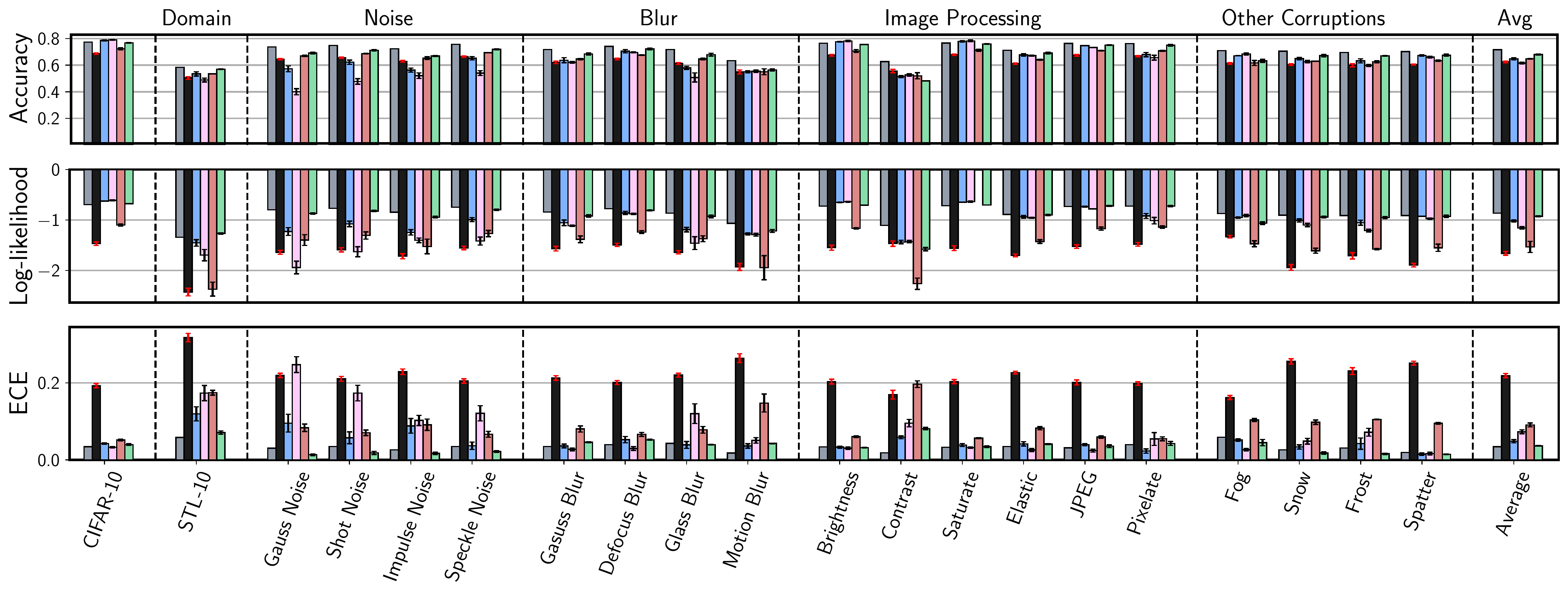}
        \caption{Corruption Intensity 3}
    \end{subfigure}
    \begin{subfigure}[t]{\textwidth}
        \includegraphics[width=0.99\textwidth]{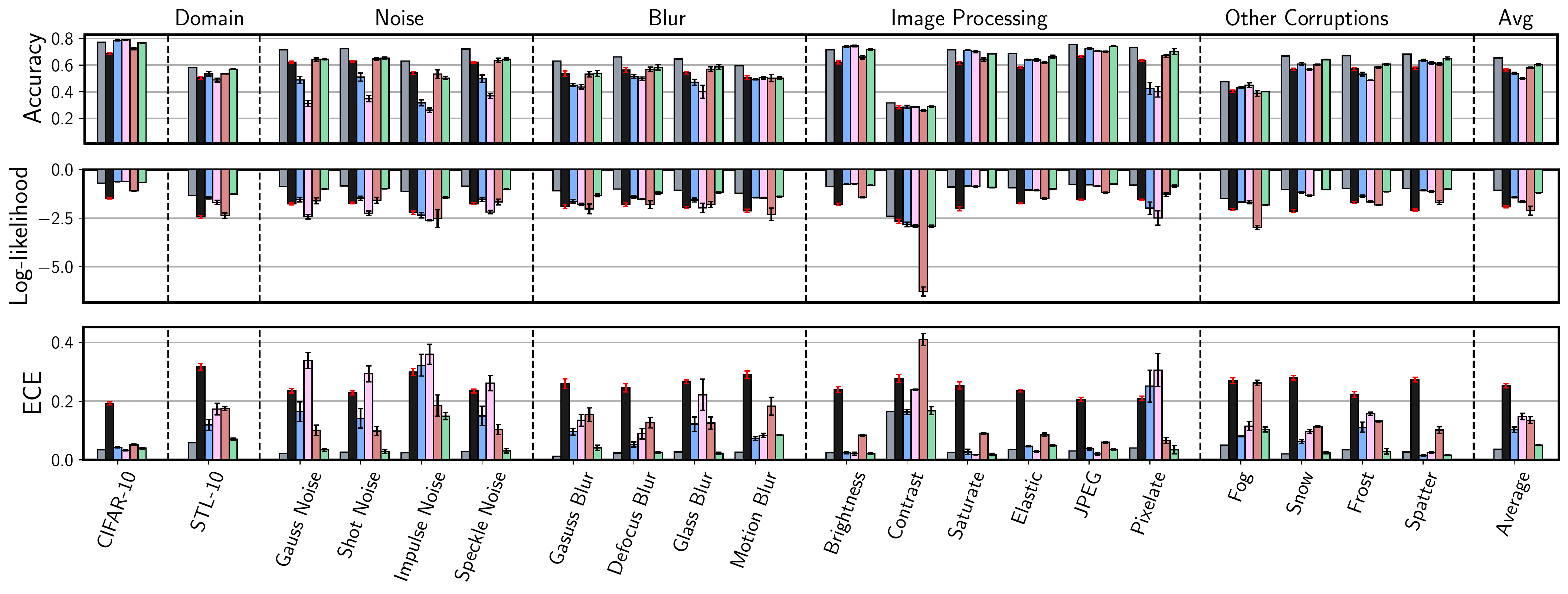}

        \quad~\includegraphics[width=0.95\textwidth]{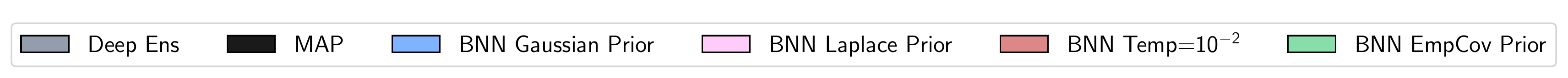}
        \caption{Corruption Intensity 5}
    \end{subfigure}
    \caption{
    \textbf{Detailed results on CIFAR-10.}
    Accuracy, log-likelihood and log-likelihood for deep ensembles, MAP solution, and BNN variants under covariate shift on CIFAR-10.
    We report the performance at corruption intensity levels 1, 3 and 5 (corruption intensity does not affect the CIFAR-10 and STL-10 columns in the plots).
    For all methods except deep ensembles we report the mean and standard deviation (via error-bars) over $3$ random seeds.
    \textit{EmpCov} priors provide the only BNN variation that consistently performs on par with deep ensembles in terms of log-likelihood and ECE.
    Tempered posteriors improve the accuracy on some of the corruptions, but significantly hurt in-domain performance.
    }
     \label{fig:detailed_results}
\end{figure}

We report the accuracy, log-likelihood and expected calibration error (ECE) for deep ensembles, MAP solutions and BMA variations in  \autoref{fig:detailed_results}.
We report the results for different corruption intensities (1, 3, 5) and provide error-bars computed over $3$ independent runs.
Across the board, \textit{EmpCov} priors provide the best performance among BNN variations on all three metrics.

\subsection{Detailed results for different priors}

\begin{figure}
    \centering
    \begin{subfigure}[t]{\textwidth}
        \includegraphics[width=0.96\textwidth]{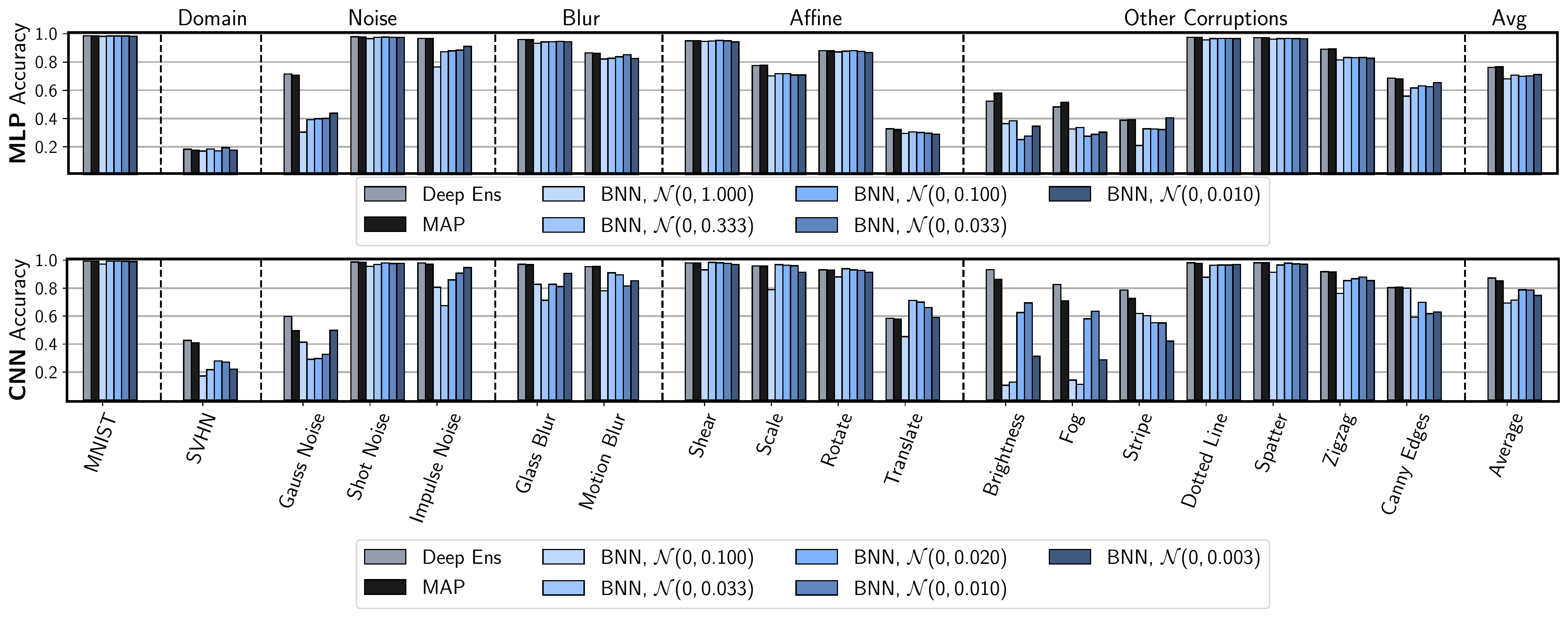}
        \caption{Gaussian priors}
    \end{subfigure}
    \begin{subfigure}[t]{\textwidth}
        \includegraphics[width=0.96\textwidth]{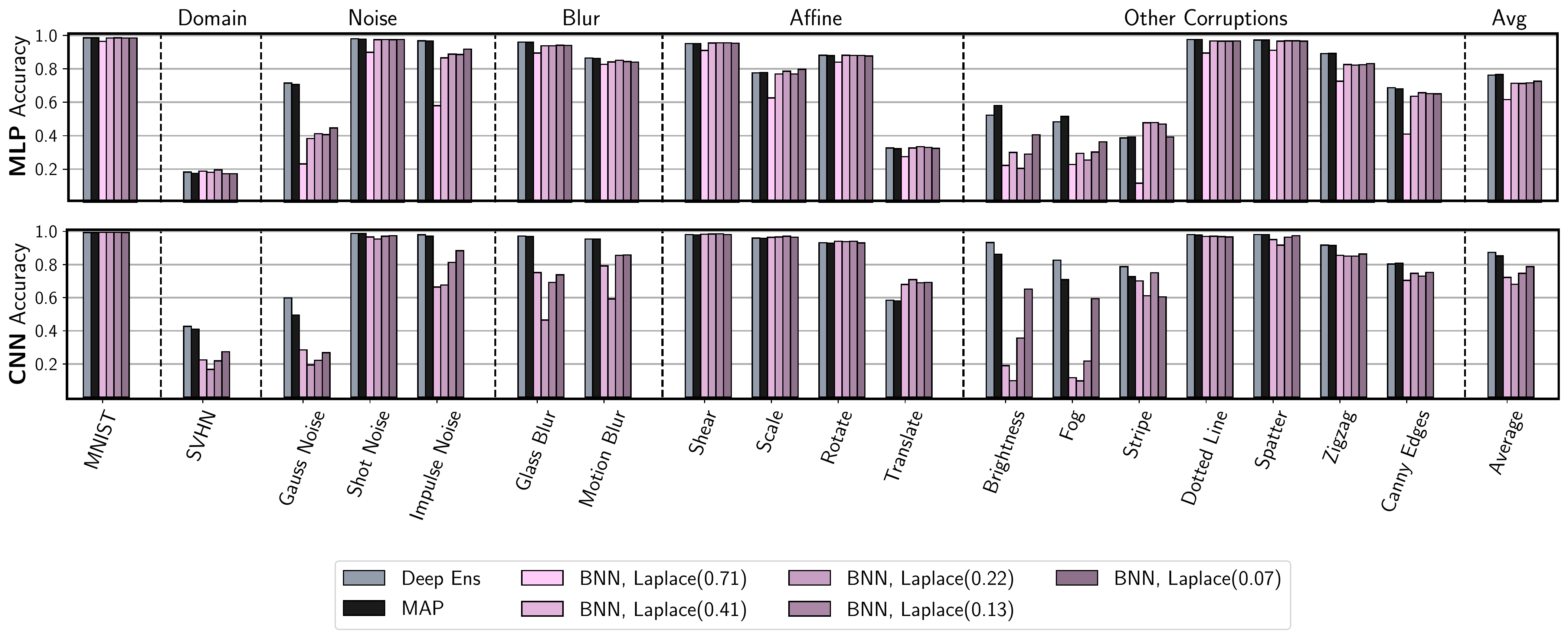}
        \caption{Laplace priors}
    \end{subfigure}
    \begin{subfigure}[t]{\textwidth}
        \includegraphics[width=0.96\textwidth]{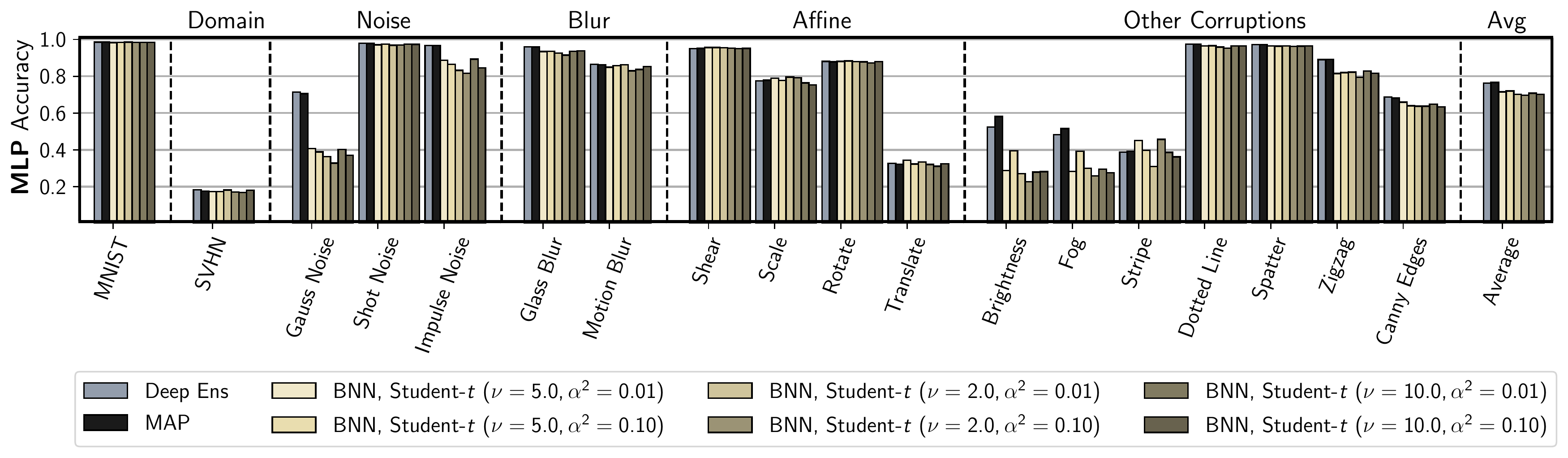}
        \caption{Student-$t$ priors}
    \end{subfigure}
    \begin{subfigure}[t]{\textwidth}
        \includegraphics[width=0.96\textwidth]{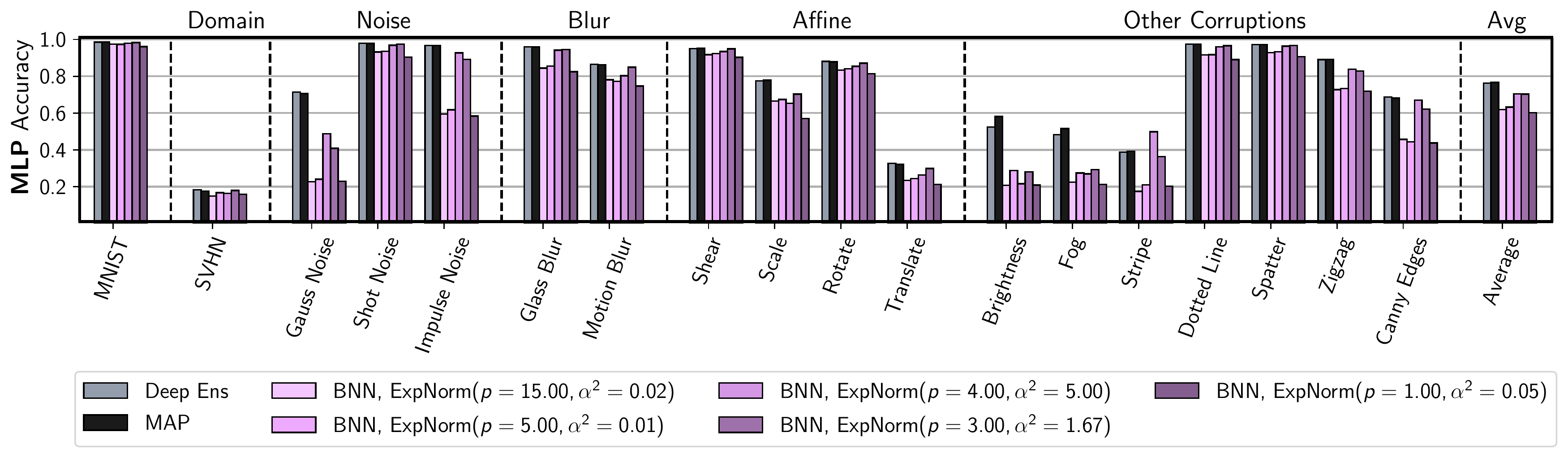}
        \caption{ExpNorm priors}
    \end{subfigure}
    \caption{
    \textbf{Priors on MNIST.}
    We report the performance of different prior families under covariate shift on MNIST.
    For Gaussian and Laplace prior families we report the results using both the MLP and CNN architectures;
    for Student-$t$ and ExpNorm we only report the results for MLP.
    None of the priors can match the MAP performance across the board, with particularly poor results under \textit{Gaussian noise}, \textit{Brightness} and \textit{Fog} corruptions.
    }
     \label{fig:app_priors}
\end{figure}

In this section, we evaluate BNNs with several prior families and provide results for different choices of hyper-parameters.
The priors are defined in \autoref{sec:app_prior_defs}.

We report the results using CNNs and MLPs on MNIST in \autoref{fig:app_priors}.
None of the considered priors completely close the gap to MAP under all corruptions. 
Gaussian priors show the worst results, losing to MAP on all MNIST-C corruptions and Gaussian noise, at all prior standard deviations.
Laplace priors show similar results to Gaussian priors under Gaussian noise, but beat MAP on the \textit{stripe} corruption in MNIST-C. 
Student-$t$ priors show better results, matching or outperforming MAP on all affine transformations, but still underpeform significantly under 
\textit{Gaussian noise}, \textit{Brightness} and \textit{Fog} corruptions.
Finally, exp-norm priors can match MAP on Shot noise and also outperform MAP on \textit{stripe}, but lose on other corruptions.
The in-domain performance with exp-norm priors is also lower compared to the other priors considered.

To sum up, none of the priors considered is able to resolve the poor robustness of BNNs under covariate shift.
In particular, all priors provide poor performance under \textit{Gaussian noise}, \textit{Brightness} and \textit{Fog} corruptions.

\section{Bayesian neural networks and spurious correlations}
\label{appendix-sec:spurious_correlations}

For corrupted data, models experience worse performance due to additional noisy features being introduced. However, it's also possible that the reverse can occur, and a seemingly highly predictive feature in the training data will not be present in the test data. This distinct category of covariate shift is often called spurious correlation. To test performance with spurious correlations, we use the Shift-MNIST dataset \citep{jacobsen2018excessive}, where we introduce spurious features via modifying the training data so that a set of ten pixels in the image perfectly correlates with class labels.

\begin{table}[h]
  \caption{\textbf{Spurious correlations.} 
  Accuracy and log-likelihood of MAP, deep ensembles and BNNs with Gaussian and \textit{EmpCov} priors on the Shift-MNIST dataset.
  }
  \label{table:shiftmnist}
  \centering
  \begin{tabular}{lllll}
    \toprule
    \cmidrule(r){1-2}
    Model     & MLP Accuracy     & MLP LL & CNN Accuracy & CNN LL \\
    \midrule
    MAP & 88.70\% & -0.527 & 48.63\% & -2.206    \\
    Deep Ensemble  & 88.73\% & -0.527 & 72.99\% & -1.041     \\
    BNN Gaussian Prior & 90.83\% & -0.598 & 64.27\% & -1.326  \\
    BNN EmpCov Prior & 86.95\% & -1.146 & 64.41\% & -1.450 \\
    \bottomrule
  \end{tabular}
\end{table}

\autoref{table:shiftmnist} shows the results for deep ensembles, MAP and BNNs with Gaussian and \textit{EmpCov} priors on the Shift-MNIST dataset. 
We see worse accuracy for CNN architectures, demonstrating how more complex architectures can more easily over-fit the spurious correlations. 
BNNs with Gaussian prior perform better than MAP on both MLP and CNN, but significantly worse than deep ensembles for CNNs.
Notice that the \textit{EmpCov} prior does not improve performance here for either architecture, highlighting the difference between spurious correlations and other forms of covariate shift. 
In particular, the largest principal components of the Shift-MNIST training dataset place large magnitude weights on the spurious features, and so using the \textit{EmpCov} prior results in samples with larger weights for the activated (spurious) pixels. 
When those same pixels are not activated in the test set, such samples will have a larger shift in their predictions. 

An in-depth analysis of BNNs in the presence of spurious correlations remains an exciting direction for further research.

\section{Bayesian neural networks in low-data regime}
\label{appendix-lowdata}

The intuition presented in Propositions \ref{prop:fc_general}, \ref{prop:conv_general} suggests that Bayesian neural networks may also underperform in low-data regime.
Indeed, if the model only observes a small number of datapoints, some of the directions in the parameter space will not be sufficiently constrained by the data.
Empirically, in Table \ref{tab:small_data} found that the performance of BNNs is indeed inferior to MAP when the training dataset is very small, but the results become more similar as the size of the dataset increases.


\begin{table}[h]
  \caption{\textbf{Spurious correlations.} 
  Accuracy of MAP and HMC BNNs using the MLP architecture on MNIST in low-data regime.
  When the dataset is very small, MAP significantly outperforms the BNN.
  }
  \label{tab:small_data}
  \centering
  \begin{tabular}{lccc}
    \toprule
    \cmidrule(r){1-2}
    & 50 datapoints & 100 datapoints & 1000 datapoints \\
    \midrule
    MAP & 66.4\%	& 74.3\%	& 90.2\%  \\
    HMC BNN	& 53.4\% & 65.4\% & 90.3\% \\
    \bottomrule
  \end{tabular}
\end{table}

We believe that the reason why we do not observe the poor generalization of the Bayesian models in the 1000 datapoints regime is that the low-variance directions are fairly consistent across the dataset.
However, in extreme low-data cases, we cannot reliably estimate the low-variance directions leading to poor performance according to Propositions \ref{prop:fc_general}, \ref{prop:conv_general}.
A detailed exploration of BNN performance in low-data regime is an exciting direction of future work.

\section{Convergence of HMC accuracy with samples}
\label{appendix-convergence}

In our experiments, we use $90$ HMC samples from the posterior to evaluate the performance of BNNs.
In this section, we verify that the Monte Carlo estimates of accuracy of the Bayesian model average converge very quickly with the number of samples, and 90 samples are sufficient for performing qualitative comparison of the methods.
In Table \ref{tab:hmc_convergence} we show the accuracy for a fully-connected HMC BNN with a Gaussian prior on MNIST under different corruptions as a function of the number of samples:

\begin{table}[h]
  \caption{\textbf{Spurious correlations.} 
  Accuracy of MAP and HMC BNNs using the MLP architecture on MNIST in low-data regime.
  When the dataset is very small, MAP significantly outperforms the BNN.
  }
  \label{tab:hmc_convergence}
  \centering
  \begin{tabular}{lccccc}
    \toprule
    \cmidrule(r){1-2}
    corruption & 10 samples & 50 samples & 100 samples & 500 samples & 1200 samples \\
    \midrule
    MNIST & 98.2\% & 98.19\% & 98.19\% & 98.32\% & 98.26\%  \\
    Impulse Noise & 85.34\% & 89.86\% & 90.68\% & 91.3\% & 91.33\%  \\
    Motion Blur & 81.56\% & 81.82\% & 82.14\% & 82.47\% & 82.61\%  \\
    Scale & 67.32\% & 68.69\% & 69.45\% & 69.91\% & 70.18\%  \\
    Brightness & 23.66\% & 20.26\% & 22.31\% & 24.08\% & 23.4\%  \\
    Stripe & 28.18\% & 30.09\% & 34.8\% & 39.26\% & 37.96\%  \\
    Canny Edges & 58.79\% & 62.85\% & 63.34\% & 64.36\% & 64.32\%  \\
    \bottomrule
  \end{tabular}
\end{table}

In each case, the performance estimated from 100 samples is very similar to the performance for 1200 samples. The slowest convergence is observed on the stripe corruption, but even there the performance at 100 samples is very predictive of the performance at 1200 samples.

\section{Tempered posteriors}
\label{appendix-tempering}

\begin{figure}
    \centering
      \includegraphics[width=0.95\textwidth]{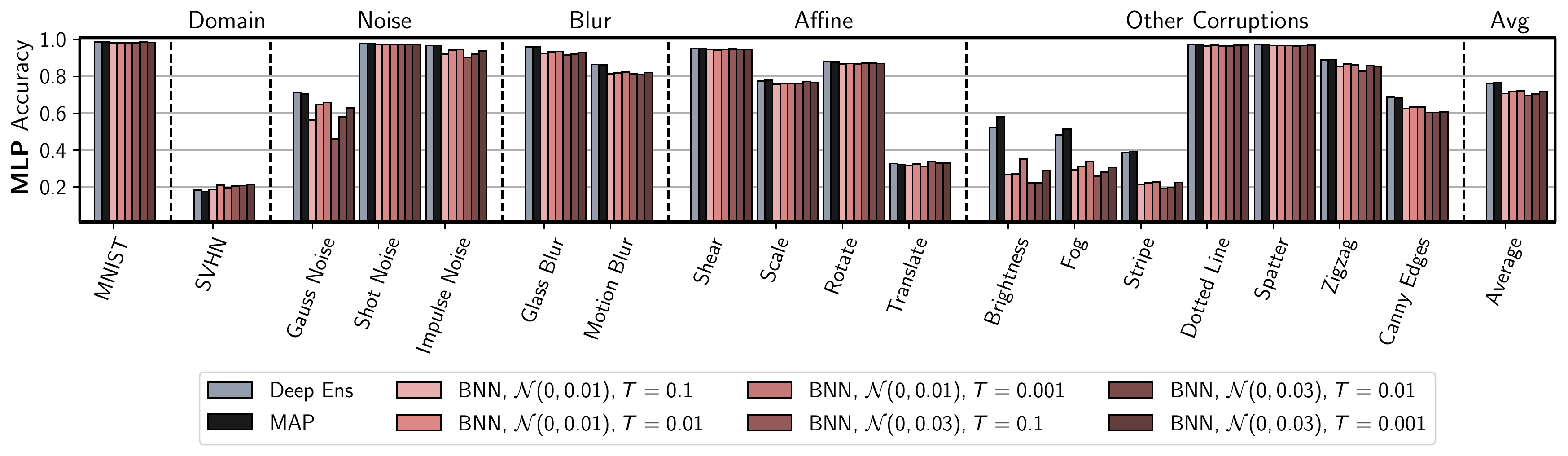}
      \caption{
      \textbf{Temperature ablation.}
      We report the performance of BNNs with Gaussian priors and tempered posteriors for different temperatures and prior scales.
      Low temperatures ($T= 10^{-2}$, $10^{-3}$) can provide a significant improvement on the \textit{noise} corruptions, but do not improve the results significantly under other corruptions.
      }
     \label{fig:app_temp_ablation}
\end{figure}

In this section we explore the effect of posterior tempering on the performance of the MLP on MNIST.
In particular, following \citet{wenzel2020good} we consider the cold posteriors:
\begin{equation}
    p_T(W \vert D) \propto
    (p(D \vert W) p(W))^{1 / T},
\end{equation}
where $T \le 1$.
In \autoref{fig:app_temp_ablation} we report the results for BNNs with Gaussian priors with variances $0.01$ and $0.03$ and posterior temperatures $T \in \{10^{-1}, 10^{-2}, 10^{-3}\}$.
As observed by \citet{izmailov2021bayesian}, lower temperatures ($10^{-2}, 10^{-3}$) improve performance under the \textit{Gaussian noise} corruption; 
however, low temperatures do not help with other corruptions significantly.

\section{Proofs of the theoretical results}
\label{appendix-sec:theory}

For convenience, in this section we assume that a constant value of $1$ is appended to the input features instead of explicitly modeling a bias vector $b$.
We assume that the output $f(x, W)$ of the network with parameters $W$ on an input $x$ is given by 
\begin{equation}
  \label{eq:parameterization}
  f(x, W) = \psi(\phi(\ldots \phi(\phi(x W^1) W^2 + b^2)) W^l  + b^{l}),
\end{equation}
where $\phi$ are non-linearities of the intermediate layers (e.g. ReLU) and $\psi$ is the final link function (e.g. softmax).

We will also assume that the likelihood is a function $\ell(\cdot, \cdot)$ that only depends on the output of the network and the target label:
\begin{equation}
  \label{eq:likelihood_parameterization}
  p(y \vert x, W) = \ell(y, f(x, W)).
\end{equation}
For example, in classification $\ell(y, f(x, W)) = f(x, W)[y]$, the component of the output of the softmax layer corresponding to the class label $y$.
Finally, we assume that the likelihood factorizes over the inputs:
\begin{equation}
  \label{eq:iid_likelihood}
  p(\mathcal D \vert W) = \prod_{x, y \in \mathcal D} p(y \vert x, W)
\end{equation}
for any collection of datapoints $\mathcal D$.

\subsection{Proof of Lemma \ref{lemma:dead_pixel}}
\label{appendix-sec:lemma1proof}

We restate the Lemma:
\begin{customlemma}{1}
\label{lemma:app_dead_pixel}
    Suppose that the input feature $x^i_k$ is equal to zero for all the examples $x_k$ in the training dataset $D$.
    Suppose the prior distribution over the parameters $p(W)$ factorizes as $p(W) = p(w^1_{ij}) \cdot p(W \setminus w^1_{ij})$ for some 
    neuron $j$ in the first layer, where $W \setminus w^1_{ij}$ represents all the parameters $W$ of the network except $w^1_{ij}$.
    Then, the posterior distribution $p(W \vert D)$ will also factorize and the marginal posterior over the parameter $w^1_{ij}$ will coincide with the prior:
    \begin{equation}
        p(W \vert D) =
        p(W \setminus w^1_{ij} \vert D) \cdot p(w^1_{ij}).
    \end{equation}
    Consequently, the MAP solution will set the weight $w^1_{ij}$ to the value with maximum prior density.
\end{customlemma}

\textbf{Proof.}\quad
Let us denote the input vector $x$ without the input feature $i$ by $x^{-i}$, and the matrix $W^1$ without
the row $i$ by $W^1_{-i}$.
We can rewrite \autoref{eq:parameterization} as follows: 
\begin{equation}
\label{eq:detailed_parameterization}
  f(x, W) = \psi(\phi(\ldots \phi(\phi(x^{-i} W^1_{-i} + \underbrace{ x^{i} W^1_{i} }_{=0} ) W^2 + b^2)) W^l + b^{l}).
\end{equation}
As for all the training inputs $x_k$ the feature $x_k^i$ is equal to $0$, the vector $x^{i} W^1_{i}$ is equal to zero and can be dropped:
\begin{equation}
  f(x_k, W) = \psi(\phi(\ldots \phi(\phi(x^{-i}_k W^1_{-i}) W^2 + b^2)) W^l + b^{l})) =: f'(x_k, W_{-i}),
\end{equation}
where $W_{-i}$ denotes the vector of parameters $W$ without $W^1_i$, and we defined a new function $f'$ that does not depend on $W_i$ and is equivalent to $f$ on the training data.
Consequently, according to \autoref{eq:likelihood_parameterization} and \autoref{eq:iid_likelihood}, we can write
\begin{equation}
  p(D \vert W) = \prod_{k=1}^n \ell(y_k, f(x_k, W)) = \prod_{k=1}^n \ell(y_k, f'(x_k, W_{-i})).
\end{equation}
In other words, the likelihood does not depend on $W^1_{i}$ and in particular $w^1_{ij}$ for any $j$.

Let us write down the posterior over the parameters using the factorization of the prior:
\begin{equation}
  \label{eq:factorized_posterior}
  p(W \vert D) = 
  \frac{\overbrace{p(D \vert W \setminus w^1_{ij}) p(W \setminus w^1_{ij})}^{\text{does not depend on $w^1_{ij}$}} p(w^1_{ij})}{Z} ,
\end{equation}
where $Z$ is a normalizing constant that does not depend on $W$.
Hence, the posterior factorizes as a product of two distributions: $p(D \vert W \setminus w^1_{ij}) p(W \setminus w^1_{ij}) / Z$ over $W \setminus w^1_{ij}$ and $p(w^1_{ij})$.
The marginal posterior over $w^1_{ij}$ thus coincides with the prior and is independent of the other parameters.

Maximizing the factorized posterior \autoref{eq:factorized_posterior} to find the MAP solution, we set the $w^1_{ij}$ to the maximum of its marginal posterior, as it is independent of the other parameters. 
$\blacksquare$

\subsection{Formal statement and proof of Proposition \ref{lemma:dead_pixel}}
\label{appendix-sec:prop1proof}

First, let us prove the following result for the MAP solution.

\begin{customproposition}{1'}
\label{prop:app_dead_pixel_map}
    Consider the following assumptions:
    \begin{enumerate}
      \item[(a)] The input feature $x^i_k$ is equal to zero for all the examples $x_k$ in the training dataset $D$.
      \item[(b)] The prior over the parameters factorizes as $p(W) = p(W_{-i}) \cdot p(W^1_i)$, where $W_{-i}$ is the vector of all parameters except for $W^1_i$, the row $i$ of the weight matrix $W^1$ of the first layer.
      \item[(c)] The prior distribution $p(W^1_{i})$ has maximum density at $0$.
    \end{enumerate}
    Consider an input $x(c) = [x^1, \ldots, x^{i-1}, c, x^{i+1}, \ldots, x^m]$.
    Then, the prediction with the MAP model $W_{MAP}$ does not depend on $c$:
        $f(x(c), W_{MAP}) = f(x(0), W_{MAP})$.
\end{customproposition}

\textbf{Proof.}\quad
Analogous to the proof of Lemma \ref{lemma:app_dead_pixel}, we can show that under the assumptions (a), (b) the posterior over the parameters factorizes as
\begin{equation}
  p(W \vert D) = p(W^{-i} \vert D) p(W^1_i).
\end{equation}
Then, the MAP solution will set the weights $W^1_i$ to the point of maximum density, which is $0$ under assumption (c).
Consequently, based on \autoref{eq:detailed_parameterization}, we can see that the output of the MAP model will not depend on $x^i = c$.
$\blacksquare$

Next, we provide results for the Bayesian model average.
We define \textit{positive-homogeneous} activations as functions $\phi$ that satisfy $\phi(c \cdot x) = c \cdot \phi(x)$ for any positive scalar $c$ and any $x$. 
For example, ReLU and Leaky ReLU activations are positive-homogeneous.

We will call a vector $z$ of class logits (inputs to softmax) $\epsilon$-\textit{separable} if the largest component $z_i$ is larger than all the other components by at least $\epsilon$:
\begin{equation}
  z_i - z_j > \epsilon \quad \forall j \ne i.
\end{equation} 

We can prove the following general proposition.

\begin{customproposition}{1''}
\label{prop:app_dead_pixel_bma}
    We will need the following assumptions:
    \begin{enumerate}
      \item[(d)] The support of the prior over the parameters $W_{-i}$ is bounded: $\|W_{-i}\| < B$.
      \item[(e)] The activations $\phi$ are positive-homogeneous and have a Lipschitz constant bounded by $L_{\phi}$.
    \end{enumerate}
    Consider an input $x(c) = [x^1, \ldots, x^{i-1}, c, x^{i+1}, \ldots, x^m]$.
    Then, we can prove the following conclusions
    \begin{enumerate}
      \item[(2)] Suppose the link function $\psi$ is identity.
        Suppose also that the expectation $\mathbb E [\phi(\ldots \phi(\phi(W^1_{i}) W^2)\ldots) W^l]$ over $W$ sampled from the posterior is non-zero.
        Then the predictive mean under BMA (see \autoref{eq:bma}) on the input $x(c)$ depends on $c$.
      \item[(3)] Suppose the link function $\psi$ is softmax.
        Then, for sufficiently large $c > 0$ the predicted class $\hat y(c) = \arg \max_{y} f(x(c), W)[y]$ does not depend on $x(c)$ for any sample $W$ from the posterior such that $z = \phi(\ldots \phi(\phi(W^1_{i}) W^2)\ldots) W^l$ is $\epsilon$-separable.
    \end{enumerate}
\end{customproposition}

\textbf{Proof.}\quad
We can rewrite \autoref{eq:parameterization} as follows: 
\begin{equation}
\begin{split}
  f(x(c), W) = \psi(\phi(\ldots \phi(\phi(x^{-i} W^1_{-i} + c W^1_{i}) W^2 + b^2)) W^l + b^{l}) =
  \\
  \psi(\phi(\ldots \phi(c \cdot \phi( [x^{-i} W^1_{-i}] / c + W^1_{i}) W^2 + b^2)) W^l + b^{l}) =
  \\
  \psi(c \cdot (\phi(\ldots \phi(\phi( [x^{-i} W^1_{-i}] / c + W^1_{i}) W^2 + b^2 / c)) W^l + b^{l} / c)).
\end{split}
\end{equation}

Now, under our assumptions the prior and hence the posterior over the weights $W_{-i}$ is bounded.
As in finite-dimensional Euclidean spaces all norms are equivalent, in particular we imply that 
(1) the spectral norms $\|W^t\|_2 < L_W$ are bounded for all layers $t = 2, \ldots, l$ by a constant $L_W$, and
(2) the Frobenious norms $\|\cdot\|$ of the bias parameters $b_t$ and the weights $W^1_{-i}$ are all bounded by a constant $B$.
We will also assume that the norm of the vector $x^{-i}$ is bounded by the same constant: $\|x^{-i}\| \le B$.

Consider the difference
\begin{equation}
\begin{split}
  \bigg\|
  & \left(\phi\left(\ldots \phi\left(\phi\left( 
    \frac{x^{-i} W^1_{-i}} {c} + W^1_{i}
  \right) W^2 +\frac{b^2}{c}\right)\right) W^l + \frac{b^{l}}{c}\right) - 
  \\
  & \phi\left(\ldots \phi\left(\phi\left( 
    \frac{x^{-i} W^1_{-i}} {c} + W^1_{i}
  \right) W^2 +\frac{b^2}{c}\right)\right) W^l
  \bigg\| 
  \le 
  \frac B c.
  \end{split}
\end{equation}
Indeed, by the $\|b^l\| \le B$.
Next, for an arbitrary $z$ we can bound
\begin{equation}
\begin{split}
  \label{eq:diff_bound}
  \left\|
  \phi\left(z + \frac {b^{l-1}}{c} \right) W^l - 
  \phi\left(z \right) W^l 
  \right\|
  \le L_W \cdot L_\phi \cdot \frac B c,
\end{split}
\end{equation}
where we used the fact that $\phi$ is Lipschitz with $L_\phi$ and the Lipschitz constant for matrix multiplication by $W^l$ coincides with the spectral norm of $W^l$ which is bounded by $L_W$.

Using the bound in \autoref{eq:diff_bound}, we have
\begin{equation}
\begin{split}
  \bigg\|
  & \left(\phi\left(\phi\left(\ldots \phi\left(\phi\left( 
    \frac{x^{-i} W^1_{-i}} {c} + W^1_{i}
  \right) W^2 +\frac{b^2}{c}\right) W^{l-1} + \frac{b^{l-1}}{c}\right)\ldots \right) W^l + \frac{b^{l}}{c}\right) - 
  \\
  & \phi\left(\phi\left(\ldots \phi\left(\phi\left( 
    \frac{x^{-i} W^1_{-i}} {c} + W^1_{i}
  \right) W^2 +\frac{b^2}{c}\right)\cdots\right)W^{l-1}\right) W^l
  \bigg\| 
  \le 
  \frac B c + L_W \cdot L_\phi \cdot \frac B c.
  \end{split}
\end{equation}
Applying the same argument to all layers of the network 
(including the first layer where $\frac{x^{-i} W^1_{-i}} {c}$ plays the role analogous to $\frac {b^{l-1}}{c}$ in \autoref{eq:diff_bound}), we get
\begin{equation}
\begin{split}
\label{eq:final_bound}
  \bigg\|
   \left(\phi\left(\dots \phi\left(\phi\left( 
    \frac{x^{-i} W^1_{-i}} {c} + W^1_{i}
  \right) W^2 +\frac{b^2}{c}\right)\dots\right) W^l + \frac{b^{l}}{c}\right)& \\
  - \phi\left(\ldots \phi\left(\phi\left( 
    W^1_{i}
  \right) W^2\right)\ldots\right) W^l &
  \bigg\| 
  \\
  \le
   \frac B c (1 + L_W \cdot L_\phi  +  L_W^2 \cdot L_\phi^2 + \ldots & + L_W^{l-1} \cdot L_\phi^{l-1}).
  \end{split}
\end{equation}
Choosing $c$ to be sufficiently large, we can make the bound in \autoref{eq:final_bound} arbitrarily tight.

\textbf{Conclusion (2)} \quad
Suppose $\psi$ is the identity. 
Then, we can write
\begin{equation}
  f(x(c), W) = c \cdot 
    \phi\left(\ldots \phi\left(\phi\left( W^1_{i}\right) W^2\right)\ldots\right) W^l
    + \Delta,
\end{equation}
where $\Delta$ is bounded: $\|\Delta\| \le B (1 + L_W \cdot L_\phi  +  L_W^2 \cdot L_\phi^2 + \ldots + L_W^{l-1} \cdot L_\phi^{l-1})$.
Consider the predictive mean under BMA,
\begin{equation}
\label{eq:conclusion2}
  \mathbb E_W f(x(c), W) = c \cdot 
    \underbrace{\mathbb E_W \phi\left(\ldots \phi\left(\phi\left( W^1_{i}\right) W^2\right)\ldots\right) W^l}_{\ne 0}
    + \underbrace{\mathbb E_W \Delta}_{\text{Bounded}},
\end{equation}
where the first term is linear in $c$ and the second term is bounded uniformly for all $c$. 
Finally, we assumed that the expectation $\mathbb E_W \phi\left(\ldots \phi\left(\phi\left( W^1_{i}\right) W^2\right)\ldots\right) W^l \ne 0$, so for large values of $c$ the first term in \autoref{eq:conclusion2} will dominate, so the output depends on $c$.

\textbf{Conclusion (3)} \quad
Now, consider the softmax link function $\psi$.
Note that for the softmax we have $\arg \max_y \psi(c \cdot z)[y] = \arg \max_y z[y]$.
In other words, multiplying the logits (inputs to the softmax) by a positive constant $c$ does not change the predicted class.
So, we have

\begin{equation}
\begin{split}
  \hat y(c, W) =
  \arg \max_y & f(x(c), W)[y] =\\
  \arg \max_y &
  \left(\phi\left(\ldots \phi\left(\phi\left( 
      \frac{x^{-i} W^1_{-i}} {c} + W^1_{i}
    \right) W^2 +\frac{b^2}{c}\right)\right) W^l + \frac{b^{l}}{c}\right)[y].
\end{split}
\end{equation}

Notice that $z_W = \phi(\ldots \phi(\phi(W^1_{i}) W^2)\ldots) W^l$ does not depend on the input $x(c)$ in any way.
Furthermore, if $z_W$ is $\epsilon$-separable, with class $y_W$  corresponding to the largest component of $z_W$, then
by taking
\begin{equation}
c > \frac {B (1 + L_W \cdot L_\phi  +  L_W^2 \cdot L_\phi^2 + \ldots + L_W^{l-1} \cdot L_\phi^{l-1})}{\epsilon},
\end{equation}
we can guarantee that the predicted class for $f(x(c), W)$ will be $y_W$ according to \autoref{eq:final_bound}.
$\blacksquare$

\subsection{General linear dependencies, Proposition \ref{prop:fc_general}}
\label{appendix-sec:linear_dependencies}

We will prove the following proposition, reducing the case of general linear dependencies to the case when an input feature is constant.

  Suppose that the prior over the weights $W^1$ in the first layer is an i.i.d. Gaussian distribution $\mathcal N(0, \alpha^2)$,
  independent of the other parameters in the model. 
  Suppose all the inputs $x_1 \ldots x_n$ in the training dataset $D$ lie in a subspace of the input space:
  $x_i^T c = 0$ for all $i = 1, \ldots, n$ and some constant vector $c$ such that $\sum_{i=1}^m c_i^2 = 1$.

Let us introduce a new basis $v_1, \ldots, v_m$ in the input space, such that the vector $c$ is the first basis vector.
We can do so e.g. by starting with the collection of vectors $\{c, e_2, \ldots, e_m\}$, where $e_i$ are the standard basis vectors in the feature space, and using the Gram–Schmidt process to orthogonalize the vectors.
We will use $V$ to denote the matrix with vectors $v_1, \ldots, v_m$ as colunms.
Due to orthogonality, we have $V V^T = I$.

We can rewrite our model from \autoref{eq:parameterization} as
\begin{equation}
  \label{eq:parameterization2}
  f(x, W) = \psi(\phi(\ldots \phi(\phi(\underbrace{x V}_{\bar x} \underbrace{V^T W^1}_{\bar W^1}) W^2 + b^2)) W^l  + b^{l}).
\end{equation}
We can thus re-parameterize the first layer of the model by using transformed inputs $\bar x = x V$, and transformed weights $\bar W^1 = V^T W^1$.
Notice that this re-parameterized model is equivalent to the original model, and doing inference in the re-parameterized model is equivalent to doing inference in the original model.

The induced prior over the weights $\bar W^1$ is $\mathcal N(0, \alpha^2 I)$, as we simply rotated the basis.
Furthermore, the input $\bar x^1_k = x^T_k v_1 = 0$ for all training inputs $k$.
Thus, with the re-parameterized model we are in the setting of Lemma \ref{lemma:dead_pixel} and Propositions  \ref{prop:app_dead_pixel_map}, \ref{prop:app_dead_pixel_bma}.

In particular, the posterior over the parameters $\bar W^1_1 = v_1^T W^1$ will coincide with the prior $\mathcal N(0, \alpha^2 I)$ (Lemma \ref{lemma:dead_pixel}).
The MAP solution will ignore the feature combination $\bar x^1 = x^T v_1$, while the BMA predictions will depend on it (Propositions  \ref{prop:app_dead_pixel_map}, \ref{prop:app_dead_pixel_bma}).

\subsection{Convolutional layers, Proposition \ref{prop:conv_general}}

Suppose that the convolutional filters in the first layer are of size $K \times K \times C$, where $C$ is the number of input channels.
Let us consider the set $\hat D$ of size $N$ of all the patches of size $K \times K \times C$ extracted from the training images in $D$ after applying the same padding as in the first convolutional layer.
Let us also denote the set of patches extracted from a fixed input image by $D_x$.

A convolutional layer applied to $x$ can be thought of as a fully-connected layer applied to all patches in $D_x$ individually, and with results concatenated:
\begin{equation}
  conv(w, x) = 
  \left\{\left(i, j, \sum_{a = i}^{i+k}\sum_{b = j}^{j+k} \sum_{c=1}^C x_{a, b, c} \cdot W^1_{a, b, c}
  \right)\right\},
\end{equation}
where $x_{a, b, c}$ is the intensity of the image at location $(a, b)$ in channel $c$, $W^1_{a, b, c}$ is the corresponding weight in the convolutional filter, and the tuples $(i, j, v)$ for all $i, j$ represent the intensities at location $(i, j)$ in the output image.

In complete analogy with Lemma \ref{lemma:dead_pixel} and Propositions \ref{prop:app_dead_pixel_map}, \ref{prop:app_dead_pixel_bma}, we can show that if all the patches in the dataset $\hat D$ are linearly dependent, then we can re-parameterize the convolutional layer so that one of the convolutional weights will always be multiplied by $0$ and will not affect the likelihood of the data.
The MAP solution will set this weight to zero, while the BMA will sample this weight from the prior, and it will affect predictions.

\section{How corruptions break linear dependence in the data}
\label{appendix-sec:corruptions_vs_pca}

\begin{figure}
    \centering
      \includegraphics[width=0.95\textwidth]{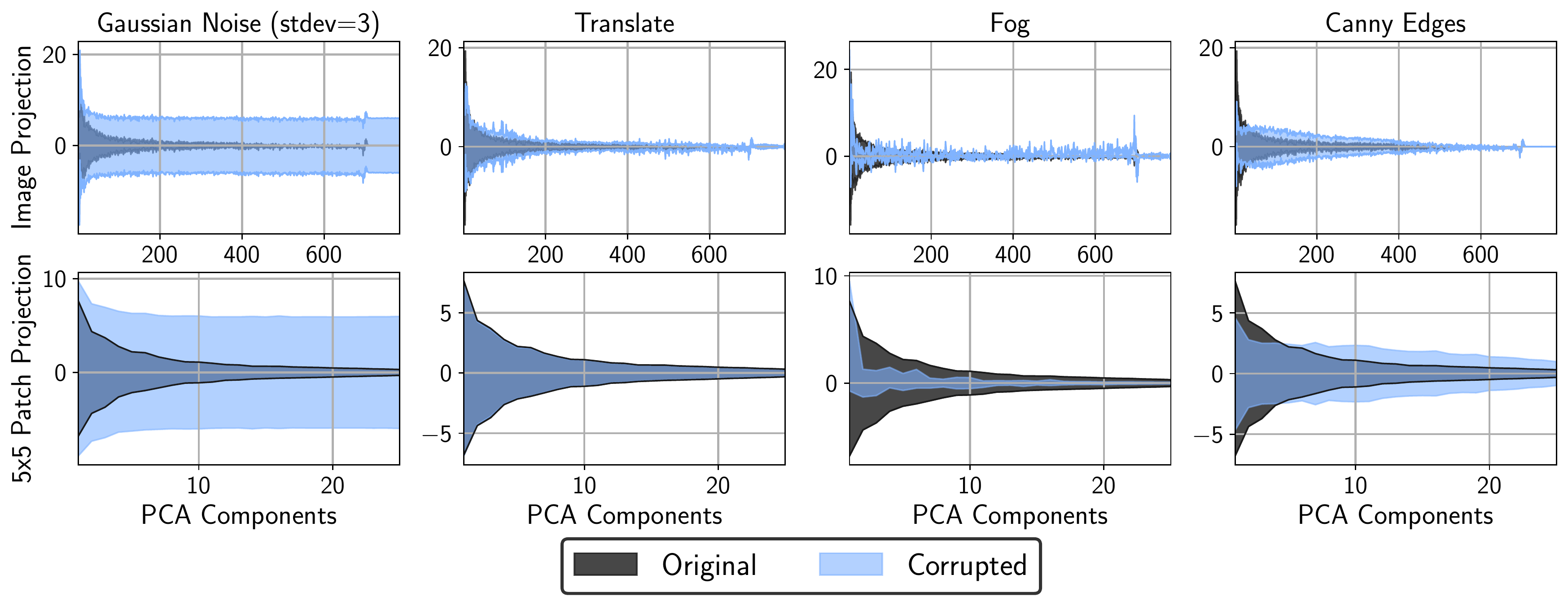}
      \caption{
      \textbf{Corruptions and linear dependence.}
      \textbf{Top}: The distribution (mean $\pm$ 2 std) of MNIST and MNIST-C images and \textbf{bottom}: $5\times5$ patches extracted from these images projected onto the corresponding principal components of the training data images and patches.
      \textit{Gaussian noise} corruption breaks linear dependencies in both cases, while \textit{Translate} does not change the projection distribution for the $5 \times 5$ patches.
      }
     \label{fig:proj_corruptions}
\end{figure}

In \autoref{fig:proj_corruptions}, we visualize the projections of the original and corrupted MNIST data on the PCA components extracted from the MNIST train set and the set of all $5 \times 5$ patches of the MNIST train set.
As we have seen in \autoref{sec:linear_dependency}, the former are important for the MLP robustness, while the latter are important for CNNs.

Certain corruptions increase variance along the lower PC directions more than others. 
For example, the \textit{Translate} corruption does not alter the principal components of the $5\times5$ patches in the images, and so a convolutional BNN with a Gaussian prior is very robust to this corruption.
In contrast, Gaussian noise increases variance similarly along all directions, breaking any linear dependencies present in the training data and resulting in much worse BNN performance. 

\section{Analyzing other approximate inference methods}
\label{appendix-sec:approximate_inference}
In this section, we provide additional discussion on why popular approximate inference methods SWAG and MC Dropout do not exhibit the same poor performance under covariate shift.

\subsection{Variational inference}

Suppose the prior is $p(w) = \mathcal N(0, \alpha^2 I)$ and the variational family contains distributions of the form $q(w) = \mathcal N(\mu, \Lambda)$, where the mean $\mu$ and the covariance matrix $\Lambda$ are parameters. Variational inference solves the following optimization problem:
maximize $\mathbb E_{w \sim \mathcal N(\mu, \Lambda)} p(D \vert w) - 
KL\left(\mathcal N(\mu, \Lambda) \vert\vert \mathcal N(0, \alpha^2 I)\right)
$ with respect to $\mu$, $\Lambda$ \citep{blundell2015weight, kingma2013auto}.

First, let us consider the case when the parameter $\Lambda$ is unconstrained and can be any positive-definite matrix.
Suppose we are using a fully-connected network, and there exists a linear dependence in the features, as in Proposition \ref{prop:fc_general}. Then, there exists a direction $d$ in the parameter space of the first layer of the model, such that the projection of the weights on this direction will not affect the likelihood, and the posterior over this projection will coincide with the prior and will be independent from other directions (Proposition \ref{prop:fc_general},), which is Gaussian. Consequently, the optimal variational distribution will match the prior in this projection, and will also be independent from the other directions, or, in other words, $d$ will be an eigenvector of the optimal $\Lambda$ with eigenvalue $\alpha^2$. So, variational inference with a general Gaussian variational family will suffer from the same exact issue that we identified for the true posterior. Furthermore, we can generalize this result to convolutional layers completely analogously to Proposition \ref{prop:conv_general},.

Now, let us consider the mean-field variational inference (MFVI) which is commonly used in practice in Bayesian deep learning. In MFVI, the covariance matrix $\Lambda$ is constrained to be diagonal. Consequently, for general linear dependencies in the features the variational distribution will not have sufficient capacity to make the posterior over the direction $d$ independent from the other directions. As a result, MFVI will not suffer as much as exact Bayesian inference from the issue presented in Propositions \ref{prop:fc_general}, \ref{prop:conv_general},.

One exception is the dead pixel scenario described in Section 5.1, where one of the features in the input is a constant zero. In this scenario, MFVI will have capacity to make the variational posterior over the corresponding weight match the prior, leading to the same lack of robustness described in Proposition \ref{prop:dead_pixel}.

\paragraph{Empirical results.}
In addition to the theoretical analysis above, we ran mean field variational inference on our fully-connected network on MNIST and evaluated robustness on the MNIST-C corruptions. Below we report the results for MFVI, MAP and HMC BNN with a Gaussian prior:

\renewcommand{\arraystretch}{1.5}
\begin{table}[h]
\small
  \caption{Accuracy of MAP, MFVI and HMC BNNs under different corruptions.
  MFVI is more robust than HMC and even outperforms the MAP solution for some of the corruptions.}
  \label{table:vi_results}
  \centering
  \begin{tabular}{lccccccc}
    \toprule
    method   & CIFAR-10 & Gaussian Noise & Motion Blur & Scale &  Brightness & Stripe & Canny Edges \\
    \midrule
    MAP & \textbf{98.5\%} & \textbf{70.6\%} & \textbf{86.7\%} & \textbf{77.6\%} & 50.6\% & 34.1\% & 68\% \\
    MFVI & 97.9\% & 62.5\% & 82.2\% & 70.5\% & \textbf{68.9\%} & \textbf{47.7\%} & \textbf{70\%} \\
    HMC & 98.2\% & 43.2\% & 82.1\% & 69.5\% & 22.3\% & 34.8\% & 63.3\% \\
    \bottomrule
  \end{tabular}
\end{table}

As expected from our theoretical analysis, MFVI is much more robust to noise than HMC BNNs.
However, on some of the corruptions (Gaussian Noise, Motion Blur, Scale) MFVI underperforms the MAP solution.
At the same time, MFVI even outperforms MAP on Brightness, Stripe and Canny Edges.

\subsection{SWAG}
SWA-Gaussian (SWAG) \citep{maddox2019simple} approximates the posterior distribution as a multivariate Gaussian with the SWA solution \citep{izmailov2018} as its mean. To construct the covariance matrix of this posterior, either the second moment (SWAG-Diagonal) or the sample covariance matrix of the SGD iterates is used. For any linear dependencies in the training data, the corresponding combinations of weights become closer to zero in later SGD iterates due to weight decay. Since SWAG only uses the last $K$ iterates in constructing its posterior, the resulting posterior will likely have very low variance in the directions of any linear dependencies. Furthermore, because SGD is often initialized at low magnitude weights, even the earlier iterates will likely have weights close to zero in these directions.

\subsection{MC Dropout}
MC Dropout applies dropout at both train and test time, thus allowing computation of model uncertainty from a single network by treating stochastic forward passes through the network as posterior samples. The full model learned at train time is still an approximate MAP solution, and thus will be minimally affected by linear dependencies in the data being broken at test time. As for the test-time dropout, we can conclude that if the expected output of the network is not affected by linear dependencies being broken, then any subset of that network (containing a subset of the network's hidden units) would be similarly unaffected. Additionally, if dropping an input breaks a linear dependency from the training data, the network (as an approximate MAP solution) is robust to such a shift.

\section{General PCA Priors}
\label{sec:general_pca_priors}

\begin{figure}
    \centering
      \includegraphics[width=0.95\textwidth]{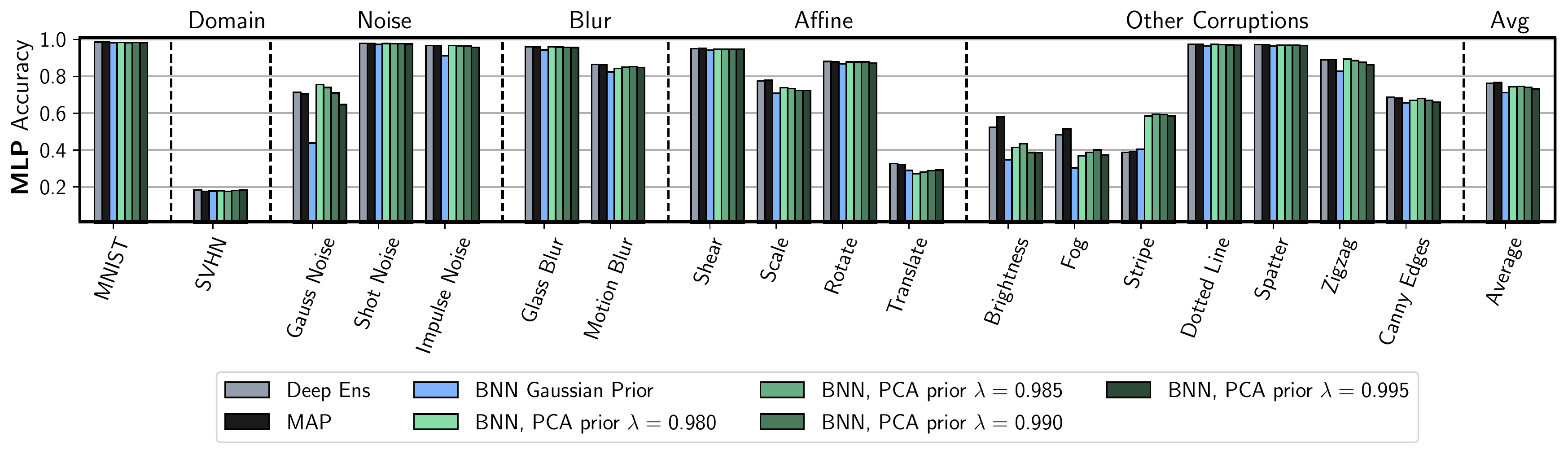}
      \caption{\textbf{General PCA priors.} 
      Performance of the PCA priors introduced in \autoref{sec:general_pca_priors} for various decay rates $\lambda$.
      PCA priors generally improve performance significantly under \textit{Gaussian noise} and \textit{Stripe}.
      Lower decay rates $\lambda$ provide better results under \textit{Gaussian noise}.
      }
     \label{fig:app_pca_general}
\end{figure}

In \autoref{sec:solution} we introduced the \textit{EmpCov} prior, which improves robustness to covariate shift by aligning with the training dataset's principal components. Following the notation used in \autoref{sec:solution}, we can define a more general family of PCA priors as

\begin{equation}
\label{eq:pca_prior}
    \begin{split}
    p(w^1) = \mathcal N(0, \alpha V\text{diag}(s)V^T + \epsilon I),\quad s_i = f(i)
    \end{split}
\end{equation}

where for an architecture with $n_w$ first layer weights, $s$ is a length $n_w$ vector, $\text{diag}(s)$ is the $n_w \times n_w$ diagonal matrix with $s$ as its diagonal, and $V$ is an $n_w \times n_w$ matrix such that the $i^{th}$ column of $V$ is the $i^{th}$ eigenvector of $\Sigma$. 

The \textit{EmpCov} prior is the PCA prior where $f(i)$ returns the $i^{th}$ eigenvalue (explained variance) of $\Sigma$. However, there might be cases where we do not want to directly use the empirical covariance, and instead use an alternate $f$. For example, in a dataset of digits written on a variety of different wallpapers, the eigenvalues for principal components corresponding to the wallpaper pattern could be much higher than those corresponding to the digit. If the task is to identify the digit, using \textit{EmpCov} might be too restrictive on digit-related features relative to wallpaper-related features.

We examine alternative PCA priors where $f(i) = \lambda^i$ for different decay rates $\lambda$. We evaluate BNNs with these priors on MNIST-C, and find that the choice of decay rate can significantly alter the performance on various corruptions. Using priors with faster decay rates (smaller $\lambda$) can provide noticeable improvement on Gaussian Noise and Zigzag corruptions, while the opposite occurs in corruptions like Translate and Fog. 
Connecting this result back to \autoref{appendix-sec:corruptions_vs_pca} and \autoref{fig:proj_corruptions}, we see that the corruptions where faster decay rates improve performance are often the ones which add more noise along the smallest principal components.

\section{Effect of non-zero mean corruptions}
\label{sec:nonzero_mean_noise}

\begin{figure}
    \centering
      \includegraphics[width=0.85\textwidth]{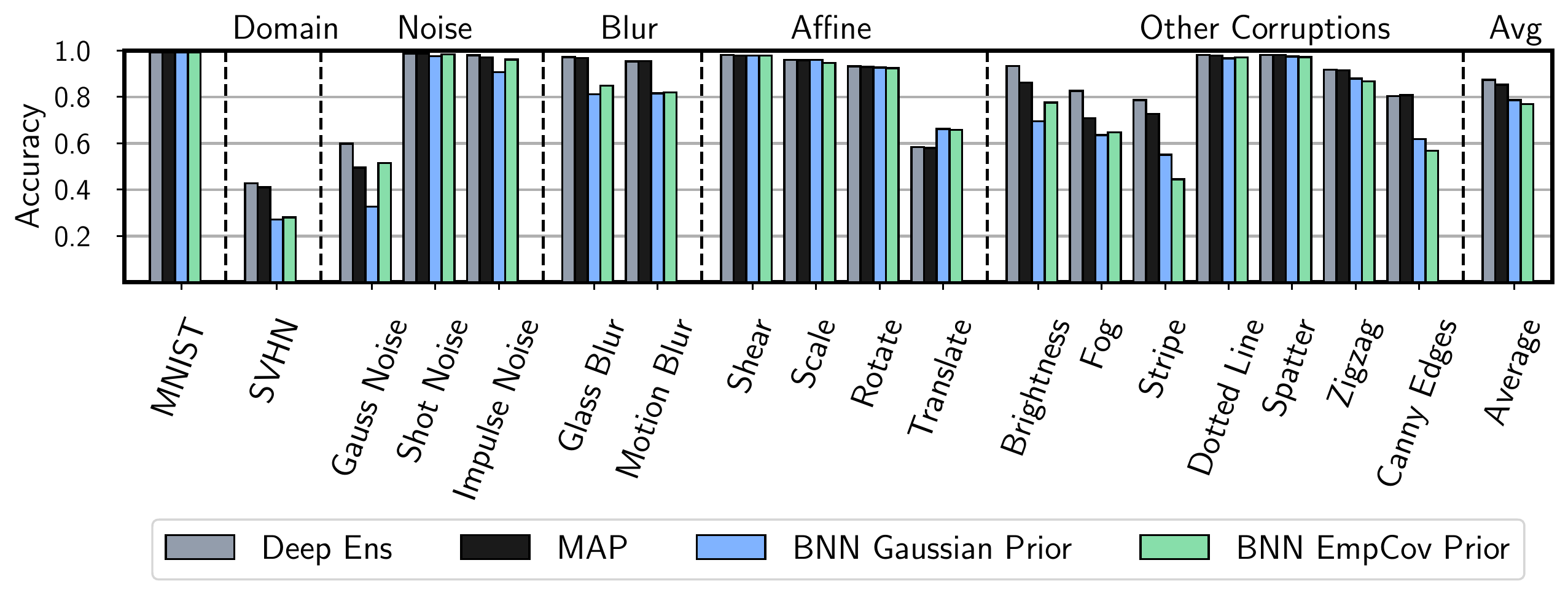}
      \caption{\textbf{MNIST CNN results.} 
      Test accuracy under covariate shift for deep ensembles, MAP optimization with SGD, and BNN with Gaussian and \textit{EmpCov} priors.
      }
     \label{fig:mnist_cnn_results}
\end{figure}


In Figure \ref{fig:mnist_cnn_results}, we report the results of deep ensembles, MAP and BNNs with Gaussian and \textit{EmpCov} priors under various corruptions using the CNN architecture on MNIST.
\textit{EmpCov} improves performance on all of the noise corruptions. For example, on the \textit{Gaussian noise} corruption, \textit{EmpCov} achieves 58.3\% accuracy while the Gaussian prior achieves 32.7\% accuracy; similarly on the \textit{impulse noise} the results are 96.5\% and 90.73\% respectively.

However, \textit{EmpCov} does not improve the results significantly on \textit{brightness} or \textit{fog}, and even hurts the performance slightly on \textit{stripe}.
Below, we explain that these corruptions are non-zero mean, and the performance is affected by the sum of the filter weights. We thus propose the SumFilter prior which greatly improves the performance on these corruptions.

\subsection{Non-zero mean corruptions}
As we have seen in various experiments (e.g. \autoref{fig:mnistc_accuracy}, \autoref{fig:app_priors}), convolutional Bayesian neural networks are particularly susceptible to the \textit{brightness}
and \textit{fog} corruptions on MNIST-C.
Both of these corruptions are not zero-mean: they shift the average value of the input features by 1.44 and 0.89 standard deviations respectively.
In order to understand why non-zero mean corruptions can be problematic, let us consider a simplified corruption that applies a constant shift $c$ to all the pixels in the input image.
Ignoring the boundary effects, the convolutional layers are linear in their input.
Denoting the output of the convolution with a filter $w$ on an input $x$ as $conv(w, x)$, and an image with all pixels equal to $1$ as $\mathbbm 1$ we can write
\begin{equation}
    conv(w, x+c \cdot \mathbbm{1}) = conv(w, x) + c \cdot conv(w, \mathbbm{1}) = 
    conv(w, x) +  \mathbbm 1 \cdot c \cdot \sum_{a,b} w_{a, b},
\end{equation}
where the last term represents an image of the same size as the output of the $conv(\cdot, \cdot)$ but with all pixels equal to the sum of the weights in the convolutional filter $w$ multiplied by $c$.
So, if the input of the convolution is shifted by a constant value $c$, the output will be shifted by a constant value $c \cdot \sum_{a,b} w_{a, b}$.

As the convolutional layer is typically followed by an activation such as ReLU, the shift in the output of the convolution can significantly hinder the performance of the network.
For example, suppose $c \cdot \sum_{a,b} w_{a, b}$ is a negative value such that all the output pixels in $ conv(w, x) +  \mathbbm 1 \cdot c \cdot \sum_{a,b} w_{a, b}$ are negative.
In this case, the output of the ReLU activation applied after the convolutional filter will be $0$ at all output locations, making it impossible to use the learned features to make predictions.

In the next section, we propose a prior that reduces the sum $\sum_{a,b} w_{a, b}$ of the filter weights, and show that it significantly improves robustness to multiple corruptions, including \textit{fog} and \textit{brightness}.

\subsection{SumFilter prior}
\label{subsec:sumfilter}

As we've discussed, if the sum of filter weights for CNNs is zero, then corrupting the input by adding a constant has no effect on our predictions. We use this insight to propose a novel prior that constrains the sum of the filter weights. More specifically, we place a Gaussian prior on the parameters and Laplace prior on the sum of the weights: 

\begin{equation}
    p(w) \sim \mathcal{N}\left(w | 0, \alpha^2 I\right) \times \text{Laplace}\Big(\sum_{\text{filter}} w | 0, \gamma^2\Big). 
\end{equation}

For our experiments, we only place the additional Laplace prior on the sum of weights in first layer filters. An alternative version could place the prior over filter sums in subsequent layers, which may be useful for deeper networks.

\subsection{Experiments}

\begin{figure}
    \centering
      \includegraphics[width=0.95\textwidth]{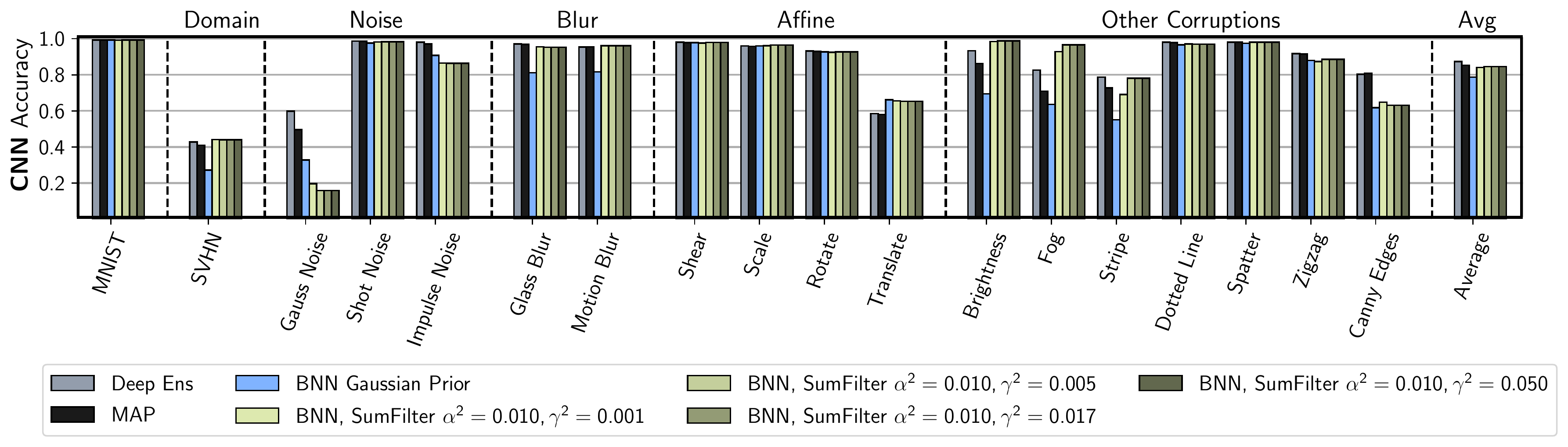}
      \caption{\textbf{SumFilter priors.}
      Performance of BNNs with the \textit{SumFilter} priors introduced in \autoref{subsec:sumfilter} for the CNN architecture on MNIST.
      \textit{SumFilter} priors do not improve the performance under \textit{Gaussian noise} unlike \textit{EmpCov priors}, but provide 
      a significant improvement on the \textit{Brightness} and \textit{Fog} corruptions.
      }
      \label{fig:app_sumfilter}
\end{figure}

\autoref{fig:app_sumfilter} shows that this prior substantially improves the performance of a convolutional BNN on MNIST-C. The BNN with a filter sum prior yields a better or comparable performance to MAP for all MNIST corruptions, with the exception of \textit{Canny Edges} and \textit{Impulse Noise}. We also implemented this prior for MLPs, but found that it only improved BNN performance on two corruptions, fog and brightness. Overall, this prior addresses a more specific issue than \textit{EmpCov}, and we would not expect it to be applicable to as many forms of covariate shift. 

\section{Example: Bayesian NALU under covariate shift}
\label{sec:bayesian_nalu}

The Neural Arithmetic Logic Unit (NALU) \citep{trask2018neural} is an architecture which can learn arithmetic functions that extrapolate to values outside those observed during training. A portion of the unit is of the form $\prod_{j=1}^m |x_j|^{w_j}$, and in this section we examine a simplified form of this unit in order to demonstrate an instance where nonlinear dependencies hurt BMA under covariate shift.

Let's consider the NALU-inspired architecture with input features $x^1, \dots, x^m$ that takes the form $f(x,w) = \prod_{j=1}^m (x^j)^{w_j}$. Suppose the prior over the weights $w = [w_1, \dots, w_m]$ is an i.i.d. Gaussian distribution $\mathcal{N}(0,\alpha^2)$. Suppose all inputs $x_1, \dots x_n$ in training dataset $\mathcal{D}$ lie in a subspace of the input space: $\prod_{j=1}^m (x_i^j)^{p_j} = 1$ for all $i = 1,\dots,n$ and some constant vector $p$ such that $\sum_{j=1}^m p_j^2 = 1$. Following the same approach as \autoref{appendix-sec:linear_dependencies}, we can introduce a new basis $v_1,\dots,v_m$ in the input space such that $v_1 = p$. We can similarly re-parameterize the model using the weights rotated into this new basis, $\bar{w} = w^Tv_1,\dots,w^Tv_m$, and it follows that $w_i = \bar{w}_1 \cdot v_1^i + \dots + \bar{w}_m \cdot v_m^i$ for all $i = 1,\dots,n$. Using the corresponding transformed inputs $\bar{x}^i = \prod_{j=1}^m (x^j)^{v_i^j}$ for all $i = 1,\dots,n$, we can rewrite our model as follows:

\begin{equation}
    \begin{split}
    f(x,w) = f(\bar{x},\bar{w}) = (\prod_{j=2}^m (\bar{x}^j)^{\bar{w}_j}) \cdot \underbrace{(\bar{x}^1)^{\bar{w}_1}}_{=1}.
    \end{split}
\end{equation}

Since $f(\bar{x},\bar{w})$ does not depend on $\bar{w}_1$ for all $\bar{x} \in \bar{\mathcal{D}}$, we can follow the same reasoning from \autoref{appendix-sec:lemma1proof} to conclude that the marginal posterior over $\bar{w}_1$ coincides with the induced prior. Since $\bar{w}$ is the result of simply rotating $w$ into a new basis, it also follows that the induced prior over $\bar{w}$ is $\mathcal{N}(0,\alpha^2I)$, and that the posterior can be factorized as $p(\bar{w}|\bar{\mathcal{D}}) = p(\bar{w}\setminus \bar{w}_1|\bar{\mathcal{D}}) \cdot p(\bar{w}_1)$.

Consider a test input $\bar{x}_k(c) = [c, \bar{x}_k^2, \dots, \bar{x}_k^m]$. The predictive mean under BMA will be:

\begin{equation}
    \begin{split}
        \mathbb E_{\bar{w}}f(\bar{x}_k(c), \bar{w}) = \mathbb E_{\bar{w}\setminus \bar{w}_1}\prod_{j=2}^m (\bar{x}_k^j)^{\bar{w}_j} \cdot \mathbb E_{\bar{w}_1}c^{\bar{w}_1}.
    \end{split}
\end{equation}

Thus the predictive mean depends upon $c$, and so the BMA will not be robust to the nonlinear dependency being broken at test time. In comparison, the MAP solution would set $\bar{w}_1=0$, and its predictions would not be affected by $c$.

While the dependency described in this section may not necessarily be common in real datasets, we highlight this example to demonstrate how a nonlinear dependency can still hurt BMA robustness. This further demonstrates how the BMA issue we've identified does not only involve \textit{linear} dependencies, but rather involves dependencies which have some relationship to the model architecture.

\section{Dead neurons}
\label{sec:dead_neurons}

Neural network models can often contain \emph{dead neurons}: hidden units which output zero for all inputs in the training set. 
This behaviour occurs in classical training when a neuron is knocked off the training data manifold, resulting in zero non-regularized gradients for the corresponding weights and thus an inability to train the neuron using the gradient signal from the non-regularized loss. However, we can envision scenarios where a significant portion of the BNN posterior distribution contains models with dead neurons, such as when using very deep, overparameterized architectures.

Let us consider the posterior distribution over the parameters $W$ conditioned on the parameters $W^1, W^2, b^2, \ldots, W^k, b^k$ of the first $k$ layers, where we use the notation of \autoref{appendix-sec:theory}.
Suppose for the parameters $W^1, W^2, b^2, \ldots, W^k, b^k$ the $k$-th layer contains a dead neuron, i.e. an output that is $0$ for all the inputs $x_j$ in the training dataset $D$.
Then, consider the sub-network containing layers $k+1, \ldots, l$.
For this sub-network, the output of a dead neuron in the $k$-th layer is an input that is $0$ for all training inputs.
We can then apply the same reasoning as we did in \autoref{appendix-sec:lemma1proof}, \autoref{appendix-sec:prop1proof} to show that there will exist a direction in the parameters $W^k$
of the $k+1$-st layer, such that along this direction the posterior \textit{conditioned} on the parameters $W^1, W^2, b^2, \ldots, W^k, b^k$ coincides with the prior (under the assumption that the prior over the parameters $W^k$ is iid and independent of the other parameters).
If a test input is corrupted in a way that activates the dead neuron, the predictive distribution of the BMA \textit{conditioned} on the parameters $W^1, W^2, b^2, \ldots, W^k, b^k$ will change.

\section{Bayesian linear regression under covariate shift}
\label{sec:app_blr}

We examine the case of Bayesian linear regression under covariate shift.  
Let us define the following Bayesian linear regression model: 

\begin{align}
y &= w^{\top} \phi(x,z) + \epsilon(x)   \\
\epsilon &\sim \mathcal{N}(0,\sigma^2)
\end{align}

where $w \in \mathbb{R}^d$ are linear weights and $z$ are the deterministic parameters of the basis function $\phi$. 
We consider the dataset $D = \{(x_i, y_i)\}_{i=1}^n$ and define 
$y := (y_1,\dots,y_N)^{\top}$, $X := (x_1, \ldots, x_n)^\top$, and $\Phi := (\phi(x_1, z), \ldots, \phi(x_n, z))^\top$

The likelihood function is given by:

\begin{equation}
    p(y| X, w, \sigma^2) = \prod_{i=1}^{n} \mathcal{N}(y_i | w^{\top} \phi(x_i,z), \sigma^2).
\end{equation}

Let us choose a conjugate prior on the weights:

\begin{equation}
    p(w) = \mathcal{N}(w | \mu_0, \Sigma_0),
\end{equation}

The posterior distribution is given by:

\begin{align*}
    p(w | \mathcal{D})  & \propto \mathcal{N}(w | \mu_0, \Sigma_0) \times \prod_{i=1}^{n} \mathcal{N}(y_i | w^{\top} \phi(x_i,z), \sigma^2) \\
    & = \mathcal{N}(w | \mu, \Sigma), \\
    \mu &= \Sigma \left(\Sigma_0^{-1} \mu_0 + \frac{1}{\sigma^2} \Phi^\top y\right), \\
    \Sigma^{-1} &= \Sigma_0^{-1} + \frac{1}{\sigma^2} \Phi^\top \Phi.
\end{align*}

The MAP solution is therefore equal to the mean,

\begin{equation}
    w_{\text{MAP}} = \Sigma \left(\Sigma_0^{-1} \mu_0 + \frac{1}{\sigma^2} \Phi^\top y\right) = \left(\Sigma_0^{-1} + \frac{1}{\sigma^2} \Phi^\top \Phi\right)^{-1}  \left(\Sigma_0^{-1} \mu_0 + \frac{1}{\sigma^2} \Phi^\top y\right) 
\end{equation}

Thus, we see that the BMA and MAP predictions coincide in Bayesian linear regression, and both will have equivalent performance under covariate shift in terms of accuracy. 

\textbf{What happens away from the data distribution?}
If the data distribution spans the entire input space, than the posterior will contract in every direction in the weight space.
However, if the data lies in a linear (or affine, if we are using a Gaussian prior) subspace of the input space,
there will be directions in the parameter space for which the posterior would coincide with the prior.
Now, if a test input does not lie in the same subspace, the predictions on that input would be affected by the shift vector according to the prior.
Specifically, if the input $x$ is shifted from the subspace containing the data by a vector $v$ orthogonal to the subspace, then the predictions between $x$ and its projection to the subspace would differ by $w^T v$, where $w \sim \mathcal N(\mu_0, \Sigma_0)$, which is itself $\mathcal N(\mu_0^T v, v^T \Sigma_0 v)$.
Assuming the prior is zero-mean, the mean of the prediction would not be affected by the shift, but the uncertainty will be highly affected.
The MAP solution on the other hand does not model uncertainty.

\section{An optimization perspective on the covariate shift problem.}
\label{sec:optimization-perspective}

In this section, we examine SGD's robustness to covariate shift from an optimization perspective. 

\subsection{Effect of regularization and initialization on SGD's robustness to covariate shift}
\label{sec:init-reg-effect}

In Section~\ref{sec:dead_neurons}, we discussed how SGD pushes the weights that correspond to dead neurons, a generalization of the dead pixels analysis, towards zero thanks to the regularization term. In this section, we study the effect of regularization and initialization on SGD's robustness under covariate shift. 

\begin{figure}
    \centering
      \includegraphics[width=0.95\textwidth]{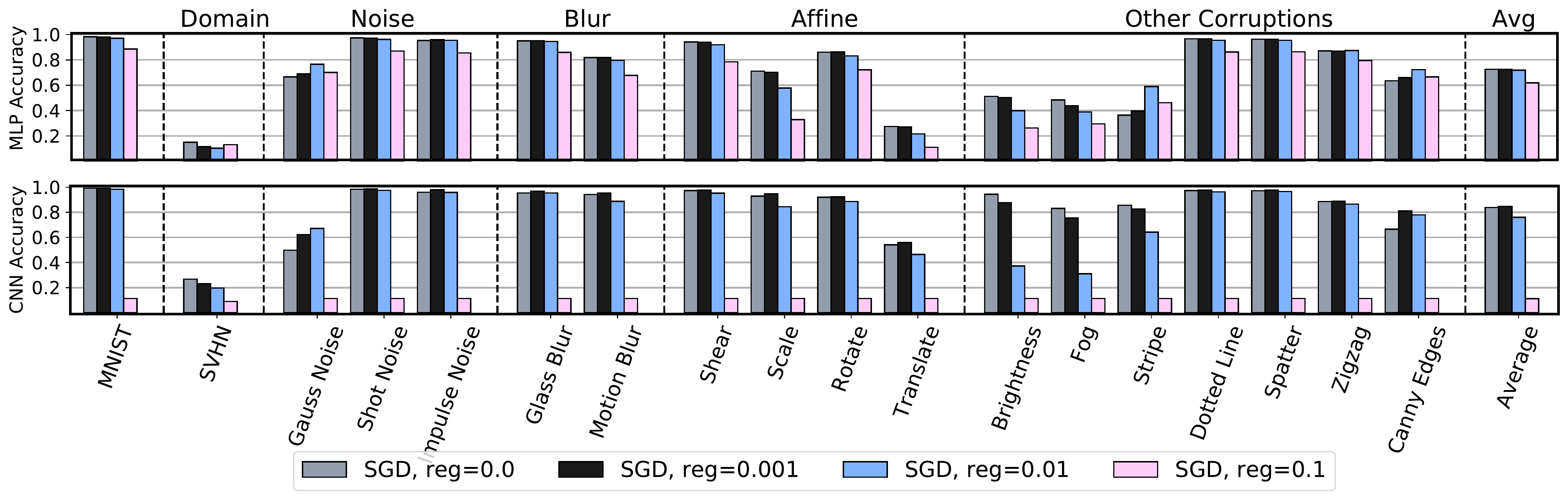}
    \caption{
    \textbf{Effect of regularization on SGD's performance on corrupted MNIST.}
    Accuracy for the following values of the regularization parameter: $0.0$, $0.001$, $0.01$, and $0.1$. 
    \textbf{Top}: Fully-connected network; \textbf{bottom}: Convolutional neural network. Regularization helps improve the performance on some corruptions, such as Gaussian noise, but its absence does not affect SGD's robustness under covariate shift because the weights are initialized at small values. 
    }
     \label{fig:effect_reg}
\end{figure}

\textbf{Regularization} \\
To study the effect of regularization on SGD's robustness under covarite shift, we hold all hyperparameters fixed and we change the value of the regularization parameter.
Figure~\ref{fig:effect_reg} shows the outcome. Most initialization schemes for neural networks initialize the weights with values close to zero, hence we expect SGD not perform as poorly on out-of-distribution data as HMC on these networks even without regularization. 
Therefore, we see that SGD without regularization (reg = $0.0$) is still competitive with reasonably regularized SGD.

\begin{figure}
    \centering
      \includegraphics[width=0.95\textwidth]{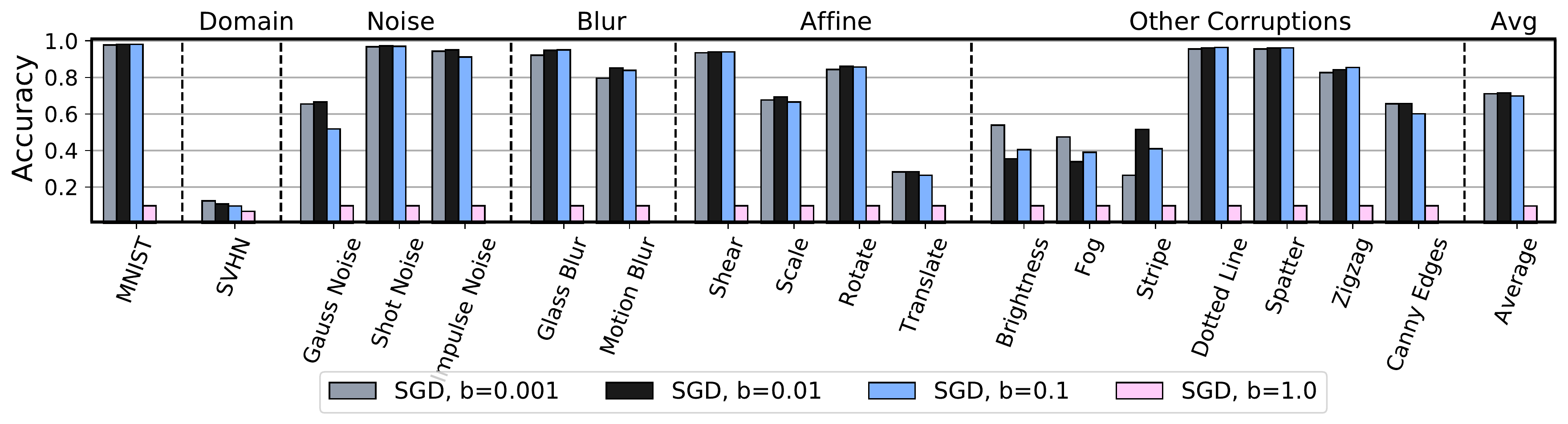}
    \caption{
    \textbf{Effect of initialization on SGD's performance on corrupted MNIST with MLP.}
    The weights are initialized using a uniform distribution  $\mathcal{U}(-b,b)$, and we consider the following values for $b$: $0.001$, $0.01$, $0.1$, and $1.0$. All experiments were run without regularization. For most corruptions, initializing the weights at smaller values leads to better robustness to covariate shift. 
    }
     \label{fig:init_exp}
\end{figure}

\textbf{Initialization} \\
The default initialization scheme for fully-connected layers in Pytorch for example is the He initialization~\citep{he2015delving}. We use a uniform initialization $\mathcal{U}(-b,b)$ and study the effect of varying $b$ on the performance of SGD under covarite shift for a fully-connected neural network. Figure~\ref{fig:init_exp} shows our empirical results, where smaller weights result in better generalization on most of the corruptions. 

\subsection{Other stochastic optimizers}
\label{sec:other-optimizers}

\begin{figure}
    \centering
      \includegraphics[width=0.95\textwidth]{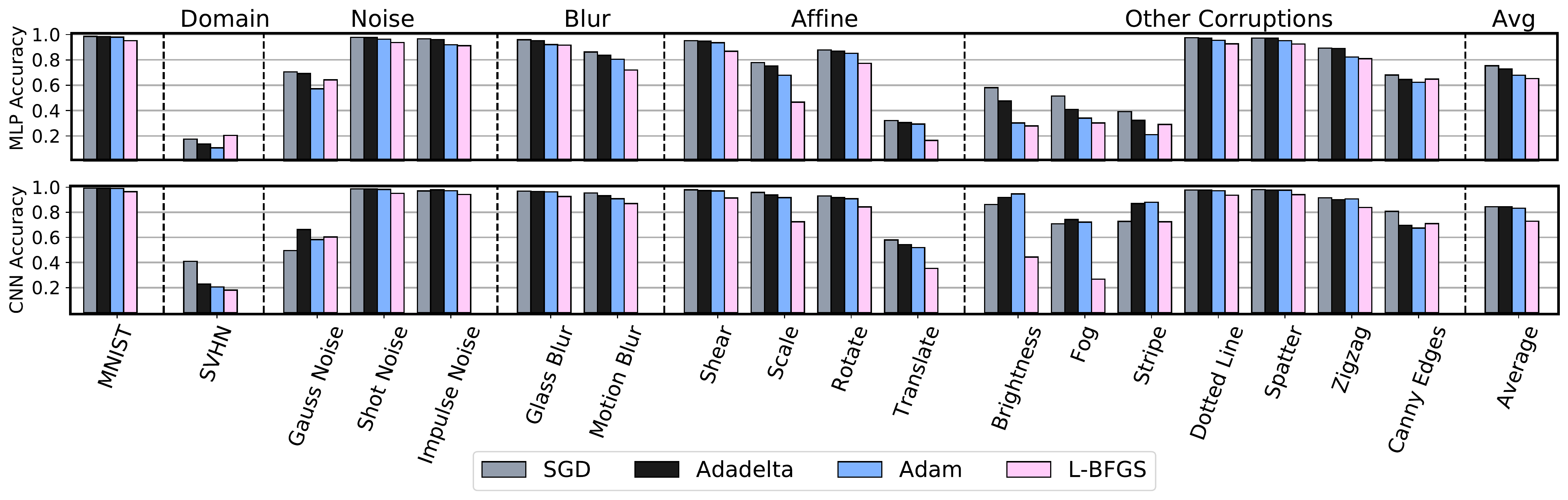}
    \caption{
    \textbf{Robustness on MNIST for different stochastic optimizers.}
    Accuracy for SGD, Adadelta, Adam and L-BFGS on MNIST under covariate shift.
    \textbf{Top}: Fully-connected network; \textbf{bottom}: Convolutional neural network.
    Adam and Adadelta provide competitive performance with SGD for most corruptions. However, SGD is better on the MLP architecture for some corruptions whereas Adam and Adadelta are better on the \textit{same} corruptions with the CNN architecture. 
    }
     \label{fig:mnistc_accuracy_optims}
\end{figure}

In addition to SGD, we examine the performance of Adam~\citep{kingma2014adam},  Adadelta~\citep{zeiler2012adadelta}, L-BFGS~\citep{nocedal1980updating, liu1989limited} on corrupted MNIST. Figure~\ref{fig:mnistc_accuracy_optims} shows the results for all 4 algorithms on the MNIST dataset under covariate shift, for both fully-connected and convolutional neural networks. We see that SGD, Adam and Adadelta have comparable performance for convolutional neural networks, whereas SGD has an edge over both algorithms on MLP. L-BFGS provides a comparatively poor performance and we hypothesise that it is due to the lack of regularization. Naive regularization of the objective function does not improve the performance of L-BFGS.

\subsection{Loss surface analysis}
\label{sec:loss-surface}

There have been several works that tried to characterize the geometric properties of the loss landscape and describe its connection to the generalization performance of neural networks. 
In particular, it is widely believed that flat minima are able to provide better generalization \citep{hochreiter1997flat, keskar2016large}. 
Intuitively, the test distribution introduces a horizontal shift in the loss landscape which makes minima that lie in flat regions of the loss surface perform well for both train and test datasets. 
From the other side, it is well-known that SGD produces flat minima. 
Hence, we would like to understand the type of distortions that corruptions in the corrupted CIFAR-10 dataset introduce in the loss surface, and evaluate the potential advantage of flat minima in this context. 

\begin{figure}
     \centering
    \begin{subfigure}[t]{0.31\textwidth}
        \raisebox{-\height}{\includegraphics[width=\textwidth]{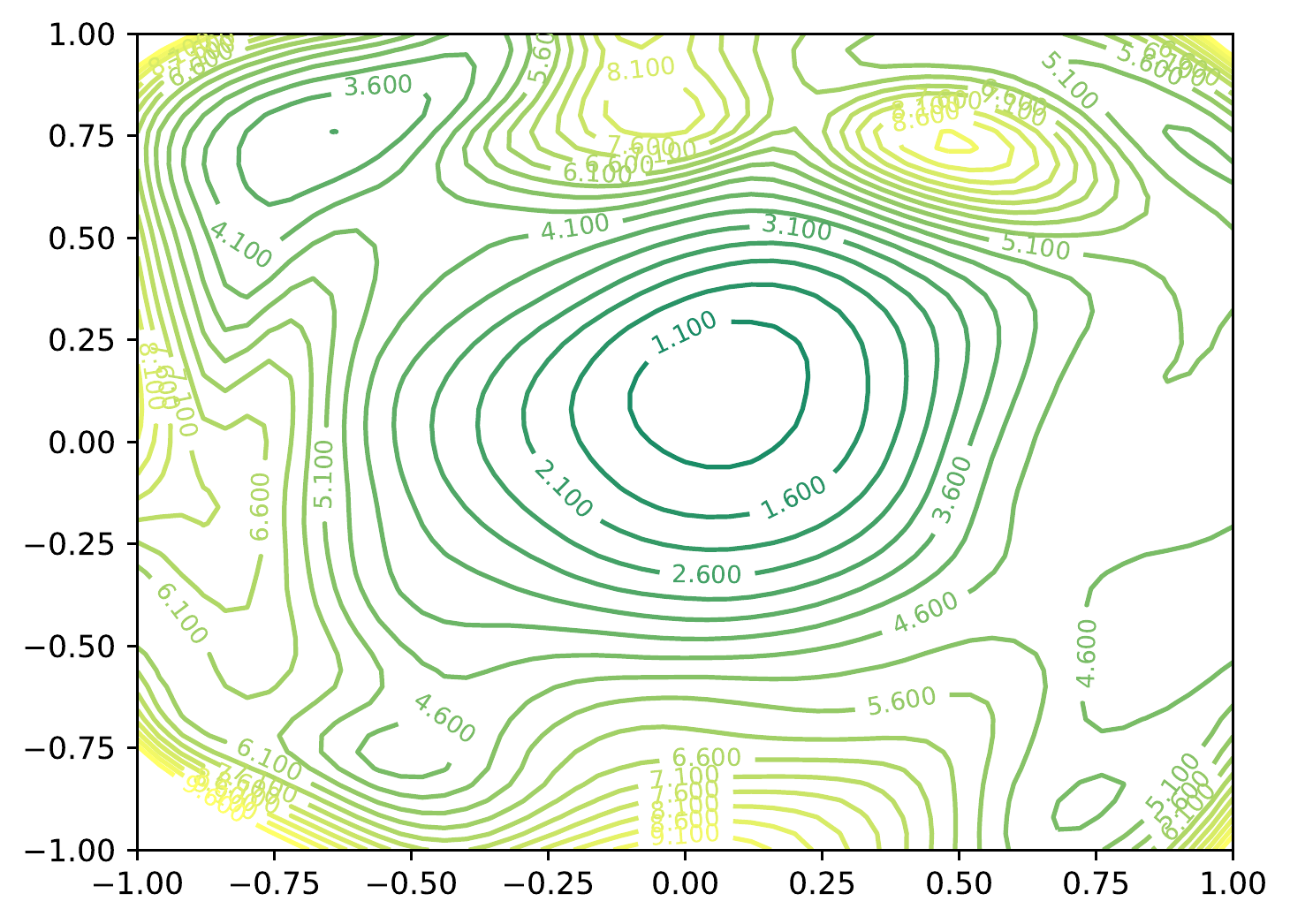}}
    \end{subfigure}
    \hfill
    \begin{subfigure}[t]{0.31\textwidth}
        \raisebox{-\height}{\includegraphics[width=\textwidth]{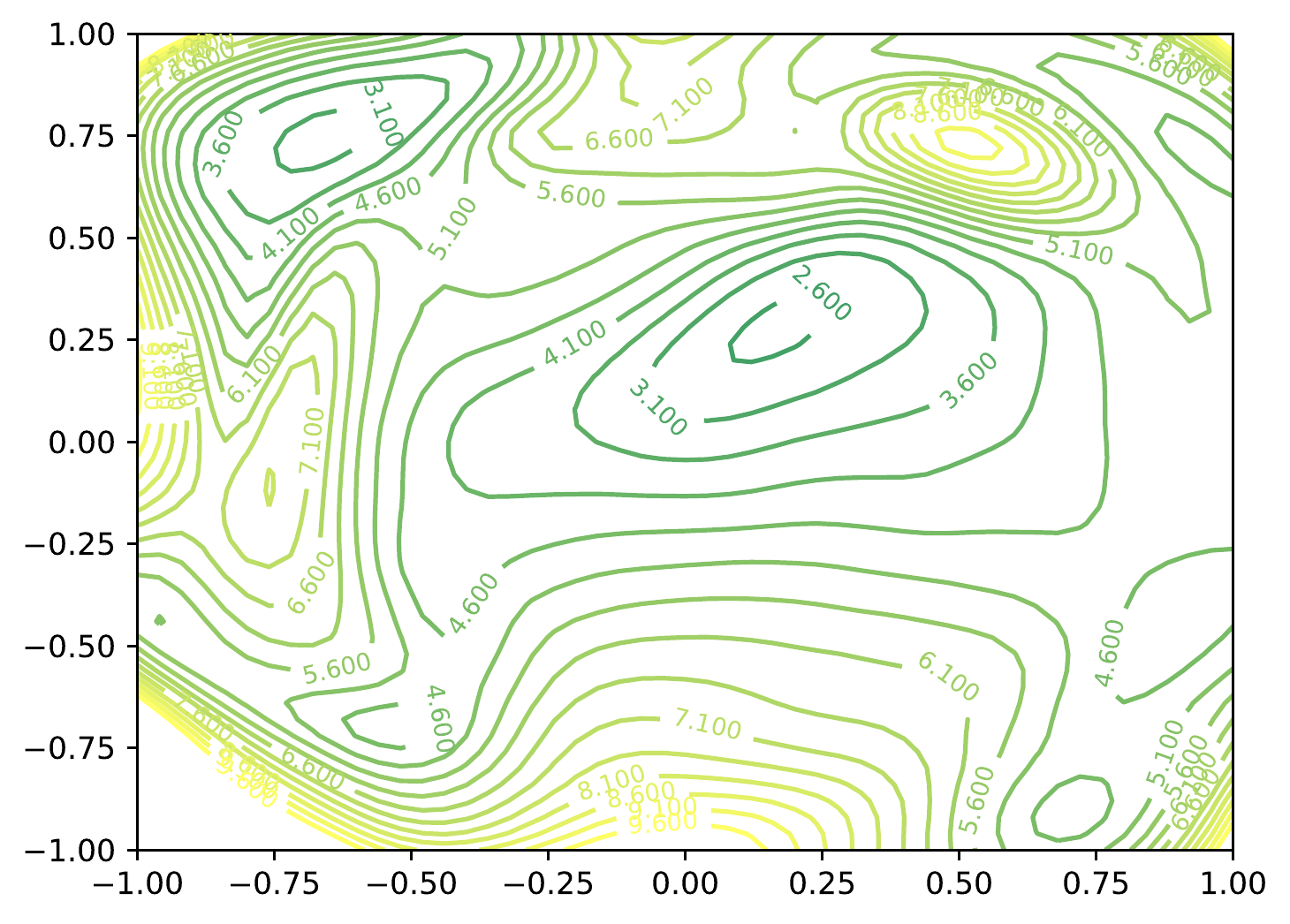}}
    \end{subfigure}
    \hfill
    \begin{subfigure}[t]{0.31\textwidth}
        \raisebox{-\height}{\includegraphics[width=\textwidth]{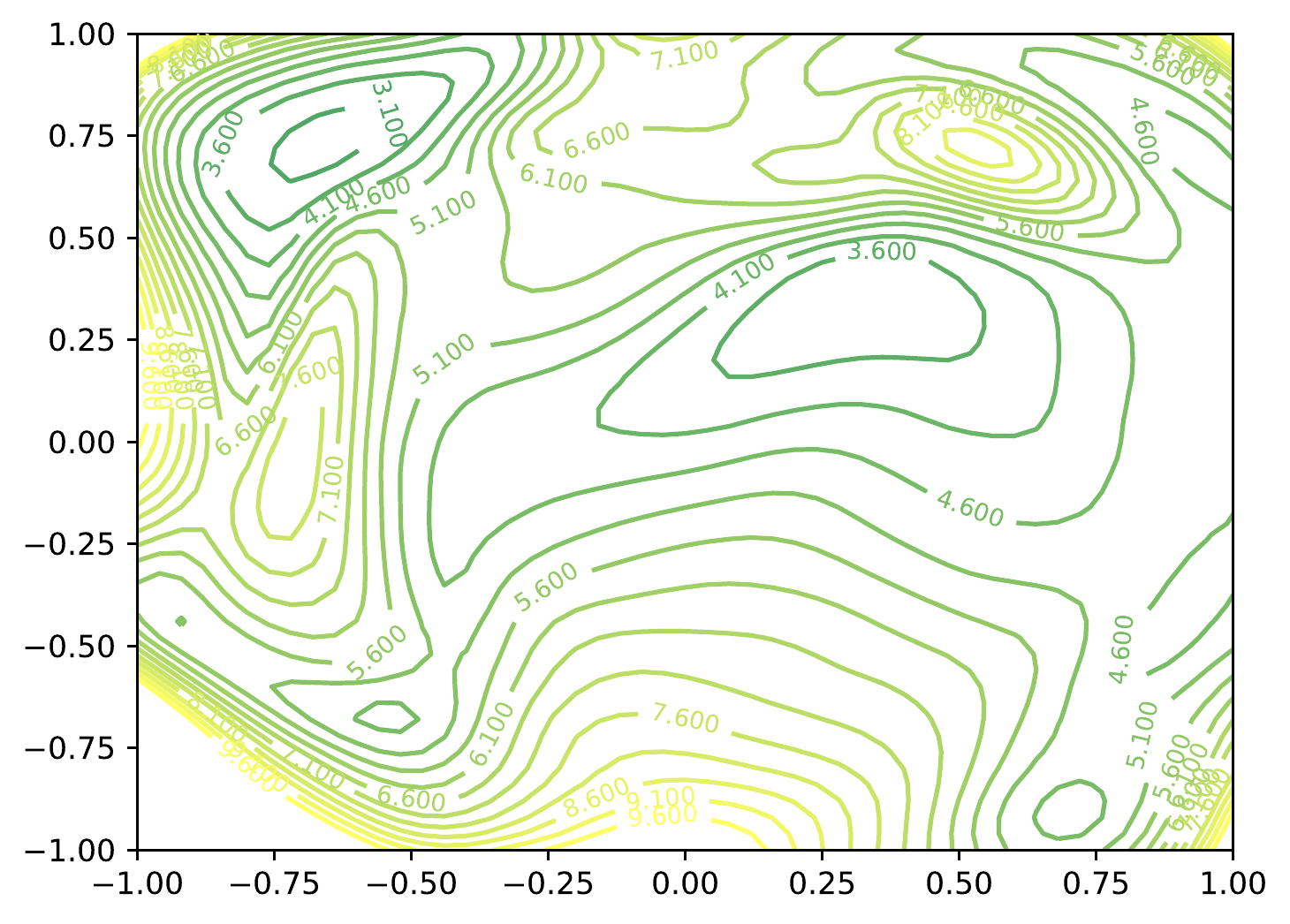}}
    \end{subfigure}
     \begin{subfigure}[t]{0.31\textwidth}
        \raisebox{-\height}{\includegraphics[width=\textwidth]{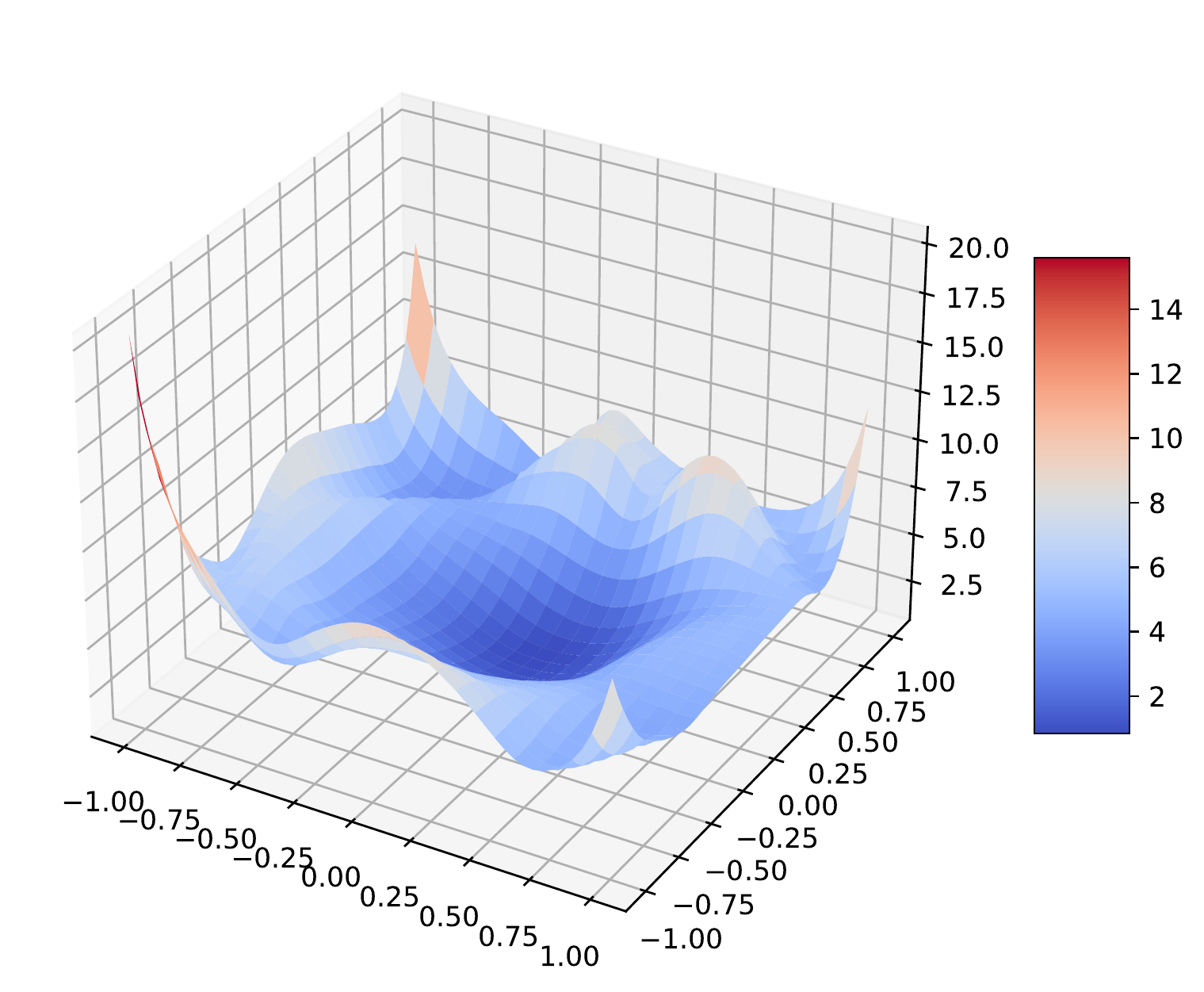}}
        \caption{Corruption intensity: 1}
    \end{subfigure}
    \hfill
    \begin{subfigure}[t]{0.31\textwidth}
        \raisebox{-\height}{\includegraphics[width=\textwidth]{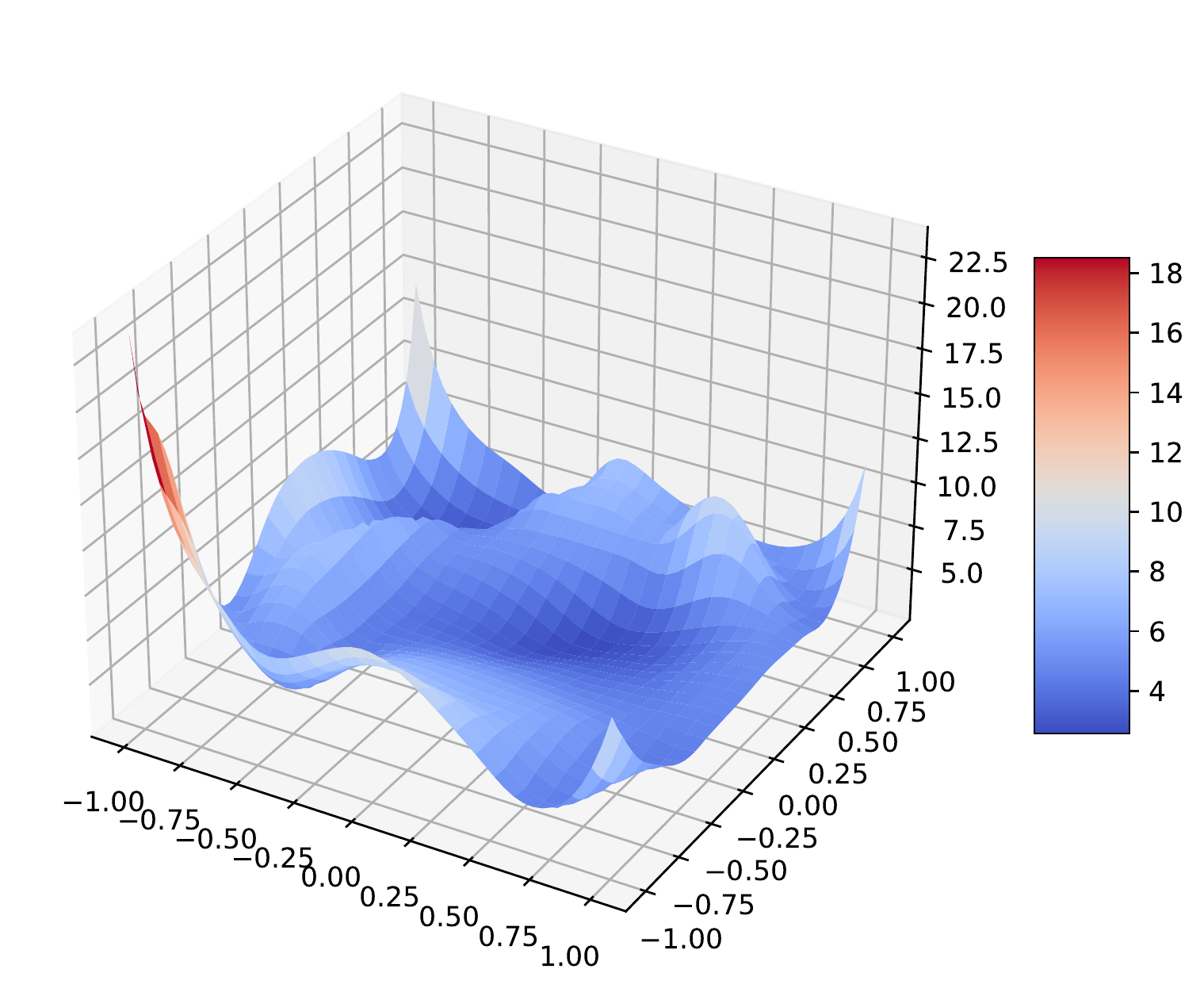}}
        \caption{Corruption intensity: 3}
    \end{subfigure}
    \hfill
    \begin{subfigure}[t]{0.31\textwidth}
        \raisebox{-\height}{\includegraphics[width=\textwidth]{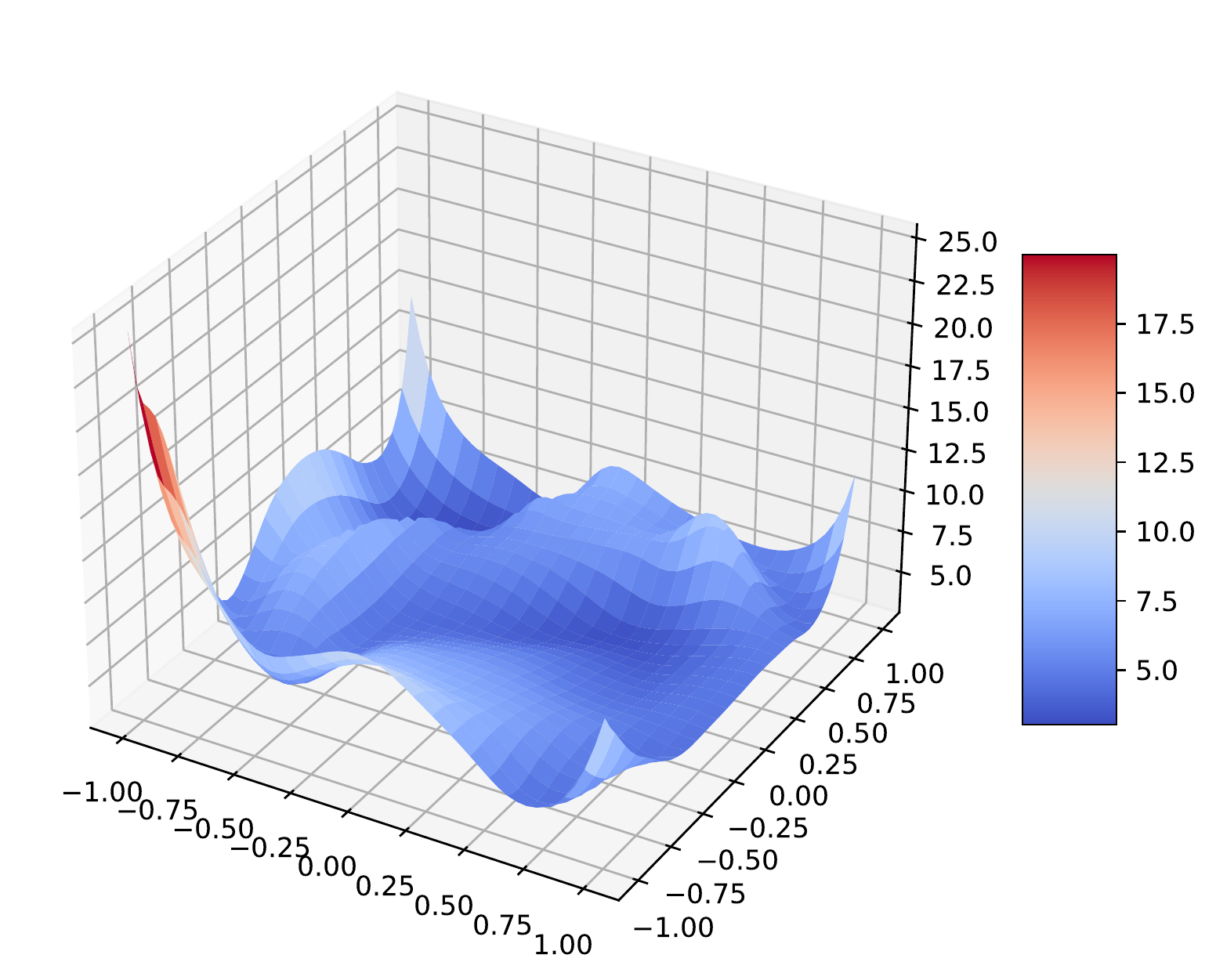}}
        \caption{Corruption intensity: 5}
    \end{subfigure}
    \caption{2D (\textbf{top}) and 3D \textbf{bottom} visualizations of the loss surface of a ResNet-56 network on corrupted CIFAR-10 with different intensity levels of the Gaussian noise corruption.}
    \label{fig:loss_vis}
\end{figure}

\begin{figure}
    \centering
    \includegraphics[scale = 0.3]{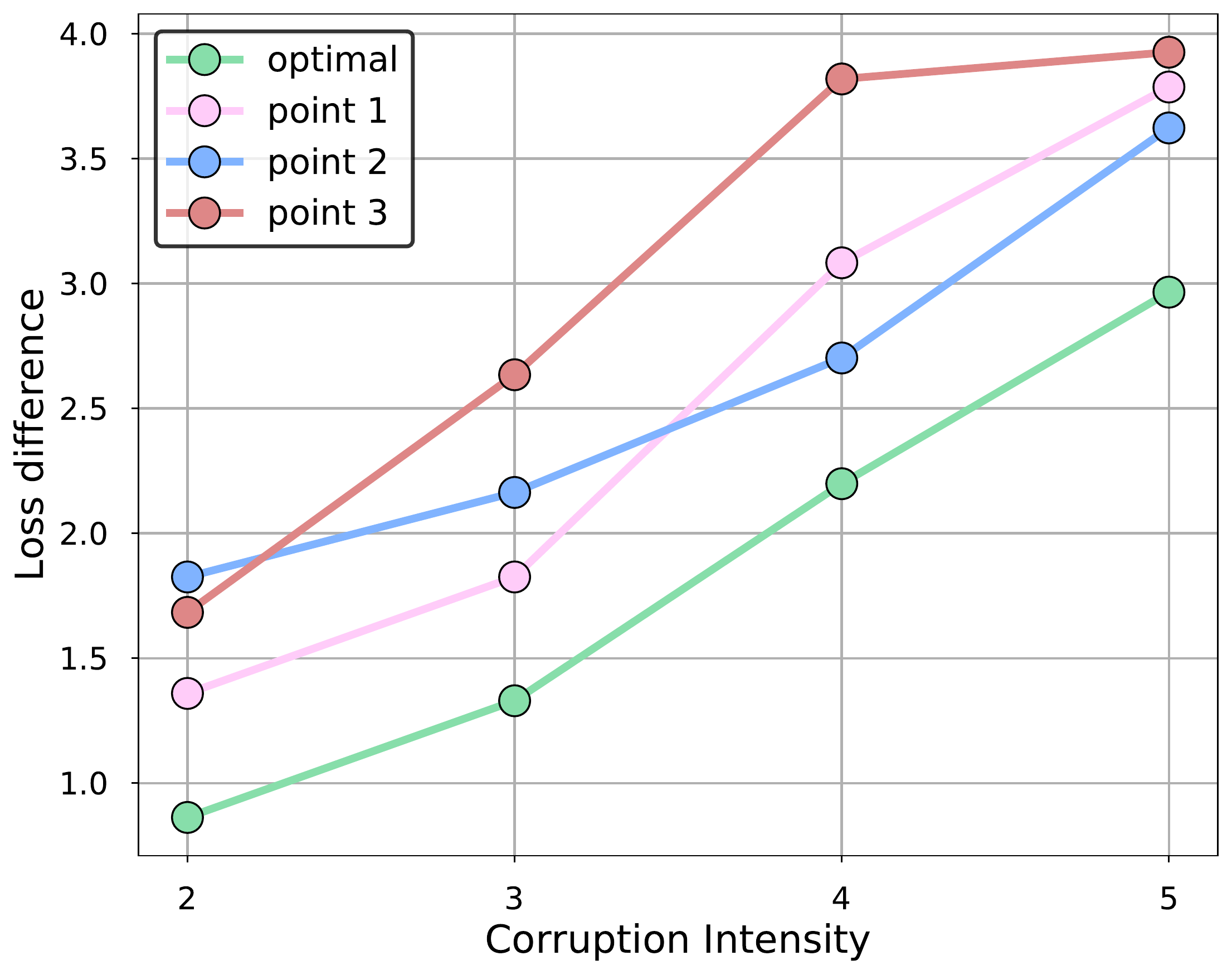}
    \caption{Loss difference between different random solutions, including the mode found through standard SGD training (\textit{optimal} in the legend), and the new mode for each corruption intensity. }\label{fig:loss_diff}
\end{figure}

In the same fashion as~\citet{li2017visualizing}, we visualize the effect of the Gaussian noise corruption on the loss surface for different intensity levels as shown in Figure~\ref{fig:loss_vis}. These plots are produced for two random directions of the parameter space for a ResNet-56 network. We observe that high levels of intensity make the loss surface more flat, but result in a worse test loss overall. We can see visually that the mode in the central flat region, that we denote $w_0$, is less affected by the corruption than a solution picked at random. Figure~\ref{fig:loss_diff} shows the loss difference between different solutions including $w_0$, that we call \textit{optimal}, and the new mode for each corruption intensity. We can see that the mode is indeed less affected by the corruptions than other randomly selected solutions of the same loss region. 

\section{Licensing}
\label{sec:app_licenses}
The MNIST dataset is made available under the terms of the Creative Commons Attribution-Share Alike 3.0 license. The CIFAR-10 dataset is made available under the MIT license. Our code is a fork of the Google Research repository at \href{https://github.com/google-research/google-research}{https://github.com/google-research/google-research}, which has source files released under the Apache 2.0 license.

\end{document}